%% file: main.tex
\documentclass[10pt]{article}
\usepackage[preprint]{tmlr}
\usepackage{graphicx}
\usepackage{amsmath,amssymb,amsfonts,mathtools,amsthm}
\usepackage{booktabs,array,tabularx,multirow,makecell}
\usepackage{nicefrac,microtype,xcolor,etoolbox}
\usepackage{turboevolve-microtype-compat}
\usepackage{algorithm,algpseudocode}
\usepackage{listings}
\usepackage{tcolorbox}
\usepackage{needspace}
\usepackage{placeins}
\usepackage{capt-of}
\usepackage[hyperfootnotes=false]{hyperref}
\usepackage{url}
\usepackage[capitalize,noabbrev]{cleveref}
\hypersetup{hidelinks,breaklinks=true,pdftitle={TurboEvolve: Towards Fast and Robust LLM-Driven Program Evolution},
  pdfauthor={Yang Yang, Zining Zhong, Jindong Li, Jiemin Wu, Kaishen Yuan, Wenshuo Chen, Menglin Yang, Yutao Yue},pdfsubject={Preprint}}
\newsavebox{\efficiencysummarybox}
\newlength{\controlcaptionwidth}

\newcommand{\best}[1]{\textbf{#1}}
\newcommand{\second}[1]{\underline{#1}}

\DeclareRobustCommand{\sysname}[1]{\textsc{#1}}
\DeclareRobustCommand{\taskname}[1]{\textnormal{\texttt{#1}}}
\newcolumntype{Y}{>{\centering\arraybackslash}X}
\newcolumntype{L}{>{\ttfamily}l}
\tcbuselibrary{listings,breakable,skins}
\lstdefinelanguage{udiff}{
  morecomment=[f][\color{red!70!black}]{-},
  morecomment=[f][\color{green!50!black}]{+},
  morecomment=[f][\color{blue!60!black}]{@@},
  morecomment=[f][\color{gray!70}]{---},
  morecomment=[f][\color{gray!70}]{+++}}
\lstdefinestyle{promptdiff}{language=udiff,basicstyle=\ttfamily,
  columns=fullflexible,keepspaces=true,showstringspaces=false,
  breaklines=true,breakatwhitespace=false,tabsize=2}
\newtcblisting{PromptDiffBox}[2][]{enhanced,breakable,colback=black!2,
  colframe=black!35,boxrule=0.35pt,arc=1.2mm,
  left=1.2mm,right=1.2mm,top=0.8mm,bottom=0.8mm,
  fonttitle=\bfseries,title={#2},listing only,
  listing options={style=promptdiff},#1}
\theoremstyle{plain}

\theoremstyle{definition}

\theoremstyle{remark}

\title{TurboEvolve: Towards Fast and Robust\\
LLM-Driven Program Evolution}
\author{%
  \name Yang Yang\textsuperscript{1}\thanks{Equal contribution.},
  Zining Zhong\textsuperscript{1}\footnotemark[1],
  Jindong Li\textsuperscript{1}, Jiemin Wu\textsuperscript{1},\\
  \name Kaishen Yuan\textsuperscript{1}, Wenshuo Chen\textsuperscript{1},
  Menglin Yang\textsuperscript{1}, Yutao Yue\textsuperscript{1,2}\thanks{Corresponding author: \texttt{yutaoyue@hkust-gz.edu.cn}. ORCID: \url{https://orcid.org/0000-0003-4532-0924}.}\\
  \addr \textsuperscript{1}The Hong Kong University of Science and Technology (Guangzhou), Guangzhou, China\\
  \textsuperscript{2}Institute of Deep Perception Technology, JITRI, Wuxi, China}

\begin{document}

\maketitle

\begin{abstract}
LLM-driven program evolution can discover high-quality programs, but high API cost and run-to-run variance hinder reliable progress.
We propose \sysname{TurboEvolve}, a multi-island evolutionary framework that improves sample efficiency and stability under matched API-cost budgets.
Inspired by multiple-offspring strategies in evolutionary algorithms, \sysname{TurboEvolve} introduces Verbalized Sampling, which elicits $K$ diverse candidates in one LLM call while using self-assigned probabilities only as generation signals, together with an online scheduler that adapts $K$ to expand exploration under stagnation and reduce redundant generation during steady progress.
To exploit existing solution pools, \sysname{TurboEvolve} further introduces island-aware initialization, which clusters seed programs into mode-specific subpopulations and applies controlled cross-cluster mixing with elitist preservation to balance diversity, sharing, and refinement.
We also curate a reusable cross-task solution-pool dataset to support warm-start initialization studies and matched-resource comparisons.
Across mathematical discovery and GPU-kernel optimization tasks, \sysname{TurboEvolve} achieves stronger cost--quality trade-offs and more stable performance than controlled baselines, while reaching competitive reported scores.
\end{abstract}

\section{Introduction}
LLM-driven program evolution is emerging as a prominent paradigm for automated program optimization and algorithm discovery \citep{lehman2023evolution,lange2024large,lange2025shinkaevolve,novikov2025alphaevolve,openevolve}. 
It combines the evolutionary loop of \emph{mutation--evaluation--selection} with LLM-based code generation: an LLM proposes candidate programs, an executable evaluator scores them, and the system iteratively retains and refines promising variants \citep{lehman2023evolution,liu2023survey}. 
This paradigm has delivered encouraging results ranging from cap-set constructions and bin packing to datacenter scheduling and compiler/hardware optimization \citep{romera2024mathematical,novikov2025alphaevolve}.

Despite this progress, practical deployments remain limited by two factors: \emph{sample efficiency} and \emph{run-to-run stability}. On challenging optimization tasks, meaningful gains may require thousands of evaluations before they materialize, making each run costly \citep{romera2024mathematical,novikov2025alphaevolve,lange2025shinkaevolve}. 
Meanwhile, stochastic evolutionary search and LLM prompt sensitivity can induce substantial run-to-run variance, making it expensive to verify whether an apparent improvement is reliable \citep{novikov2025alphaevolve,lange2025shinkaevolve}. 
These constraints motivate LLM-driven program evolution that is both more efficient and more robust under the same budget.

To this end, we build \sysname{TurboEvolve} on \sysname{OpenEvolve}'s multi-island,
archive-based program-evolution backbone~\citep{openevolve}.
Its guiding principle is to increase the \emph{effective information yield per LLM call}:
instead of acquiring candidates one-by-one through repeated invocations, \sysname{TurboEvolve} generates complementary candidate sets within one context, reducing dependence on any single stochastic rewrite~\citep{zhang2025verbalized}.
It further adapts the per-round candidate count online, allocating generation budget between exploration and exploitation based on evolutionary feedback~\citep{hevia2021self}.

Efficient program evolution can also benefit from reusable initialization. 
In many realistic settings, prior runs or external sources provide a pool of candidate programs, often including strong solutions; reusing them can accelerate search and reduce wasted budget \citep{kazimipour2014review}. 
However, existing frameworks typically support only a single seed, a generic initial pool, or manual injection \citep{romera2024mathematical,novikov2025alphaevolve,openevolve,assumpccao2025codeevolve,lange2025shinkaevolve}, without treating \emph{island-aware allocation of an external solution pool} as a first-class problem. 
\sysname{TurboEvolve} therefore introduces an island-aware initialization strategy: it embeds and clusters candidate programs into mode-specific subpopulations \citep{guo2022unixcoder}, then initializes islands with controlled cross-cluster mixing and elitist preservation. 
To support this setting, we further curate a cross-task solution-pool dataset\footnote{The curated solution-pool dataset is available at \url{https://anonymous.4open.science/r/dataset-B8DC}.} that aggregates high-scoring candidate programs and metadata across tasks. 
Unlike prior releases that mainly provide evaluators or a small set of examples, this dataset serves as a reusable warm-start resource for initialization studies, reproducible comparisons under matched resources, and future work on candidate retrieval or prompt construction \citep{fernando2023promptbreeder,wan2025loongflow}.

In summary, \sysname{TurboEvolve} improves sample efficiency and run-to-run stability for LLM-driven program evolution under costly model calls and executable evaluations. 
Our contributions are fourfold. \textbf{(1)} We reinterpret Verbalized Sampling as a multi-offspring operator, eliciting $K$ diverse candidates per LLM call while using self-assigned probabilities only as generation signals. \textbf{(2)} We propose an island-wise adaptive $K$ scheduler that adjusts search width online, expanding under stagnation and reducing redundant generation during sustained progress. \textbf{(3)} We introduce island-aware warm-start initialization from external solution pools via representation-based clustering, cross-cluster mixing, and elitist preservation, and curate a reusable cross-task solution-pool dataset for initialization studies and matched-resource comparisons. \textbf{(4)} Across mathematical discovery and GPU-kernel optimization tasks, \sysname{TurboEvolve} achieves stronger cost--quality trade-offs and more stable performance under matched budgets.

\section{Related Work}

\subsection{Evolutionary Computation: Dynamic Search Control and Island-Aware Initialization}

Evolutionary computation has long studied candidate generation, selection and retention, parameter control, and population initialization.
Multiple-offspring or brood-size strategies enlarge local search breadth by producing several descendants in one evolutionary step \citep{zhang2006gp,latorre2008using};
dynamic parameter control adjusts mutation rates, population sizes, or search intensity according to the search state \citep{hevia2021self,doerr2019theory};
and island models maintain diversity through subpopulation separation and migration \citep{cantu1998survey,mouret2015illuminating}.
Initialization and warm-starting also affect convergence and diversity \citep{kazimipour2014review}, while niching methods such as fitness sharing emphasize preserving multiple search modes \citep{goldberg1987genetic}.
However, these mechanisms are typically studied with explicit variation operators, cheap offspring generation, and initialization schemes based on random populations, a single initial solution, a few manually designed seeds, or generic seeding strategies.
In LLM-driven program evolution, these design axes become cost-sensitive: the number of candidates per model call becomes a search-width control variable, and redundant initialization wastes expensive generation and evaluation budget.
Moreover, when a heterogeneous pool of strong candidate programs is available, it remains underexplored how such seed programs should be separated into modes, assigned across islands, selectively shared, and protected.
\sysname{TurboEvolve} reformulates these classical levers for the LLM setting through budget-aware candidate-count control and island-aware external-pool allocation.

\Needspace{4\baselineskip}
\subsection{LLM-Driven Program Evolution and Automatic Heuristic Design}
Recent LLM-driven program evolution is largely organized around executable-feedback loops, where LLMs propose candidate programs and automatic evaluators retain promising variants.
Systems such as \sysname{FunSearch}~\citep{romera2024mathematical} and \sysname{AlphaEvolve}~\citep{novikov2025alphaevolve} demonstrate this paradigm for mathematical discovery, algorithm design, and program optimization, while open-source frameworks further develop archive-based evolutionary search in modular or sample-efficient forms.
Examples include \sysname{OpenEvolve}~\mbox{\citep{openevolve}}, \sysname{CodeEvolve}~\mbox{\citep{assumpccao2025codeevolve}},
\sysname{ShinkaEvolve}~\mbox{\citep{lange2025shinkaevolve}}, and \sysname{GigaEvo}~\mbox{\citep{khrulkov2025gigaevo}}.
In automatic heuristic design, one line of work improves the construction of individual offspring, for example by evolving natural-language ideas with code or using reflection to refine generation~\citep{liu2024evolution,ye2024reevo};
another line strengthens the global search process through tree search, partitioning, niching, or explicit resource allocation~\citep{zheng2025monte,hu2025partition}.
Recent systems also explore richer planning, memory, task-specific control, large-neighborhood search, or test-time learning mechanisms~\citep{wan2025loongflow,ye2025large,wang2025thetaevolve,yuksekgonul2026learning}.
\sysname{TurboEvolve} targets a complementary design axis: instead of redesigning the planner or search controller, it adds lightweight system-level controls to an archive-based, island-style backbone.
Specifically, it improves per-call candidate yield, schedules the candidate count online using evolutionary feedback, and injects external solution pools into multi-island search in a structured way.
Our solution-pool dataset provides reusable warm-start resources for initialization studies and matched-budget comparisons.

\subsection{Verbalized Sampling for Budget-Aware Multi-Offspring Control}

Verbalized Sampling (VS) is an inference-time diversification method that asks an LLM to generate multiple distinct candidates in one call and verbalize a self-assigned probability or weight for each candidate, mitigating output collapse and broadening the response distribution \citep{zhang2025verbalized}. Prior work mainly studies VS as a prompting-level diversity mechanism, without addressing how its candidates interact with executable feedback, archive updates, or fixed search budgets. In \sysname{TurboEvolve}, we reinterpret VS as a multi-offspring operator for LLM-driven program evolution: a single model invocation produces a set of candidate programs under the same evolutionary context, while selection remains evaluator- and archive-driven. Thus, the verbalized weights serve as elicitation signals rather than downstream selection scores. This turns VS from a static diversity prompt into a budget-aware proposal mechanism that improves per-call candidate yield and amortizes shared prompt overhead. Beyond using a fixed number of VS candidates, \sysname{TurboEvolve} further treats the candidate count $K$ as a search-width control variable, inspired by dynamic parameter control and search-width adaptation in evolutionary computation \citep{hevia2021self,doerr2019theory}. By increasing $K$ under stagnation and decreasing it during sustained progress, \sysname{TurboEvolve} couples inference-time diversification with evolutionary feedback, yielding dynamic multi-offspring control for archive-based evolution.

\section{Method}

\begin{figure}[t]
    \centering
    \includegraphics[width=\linewidth]{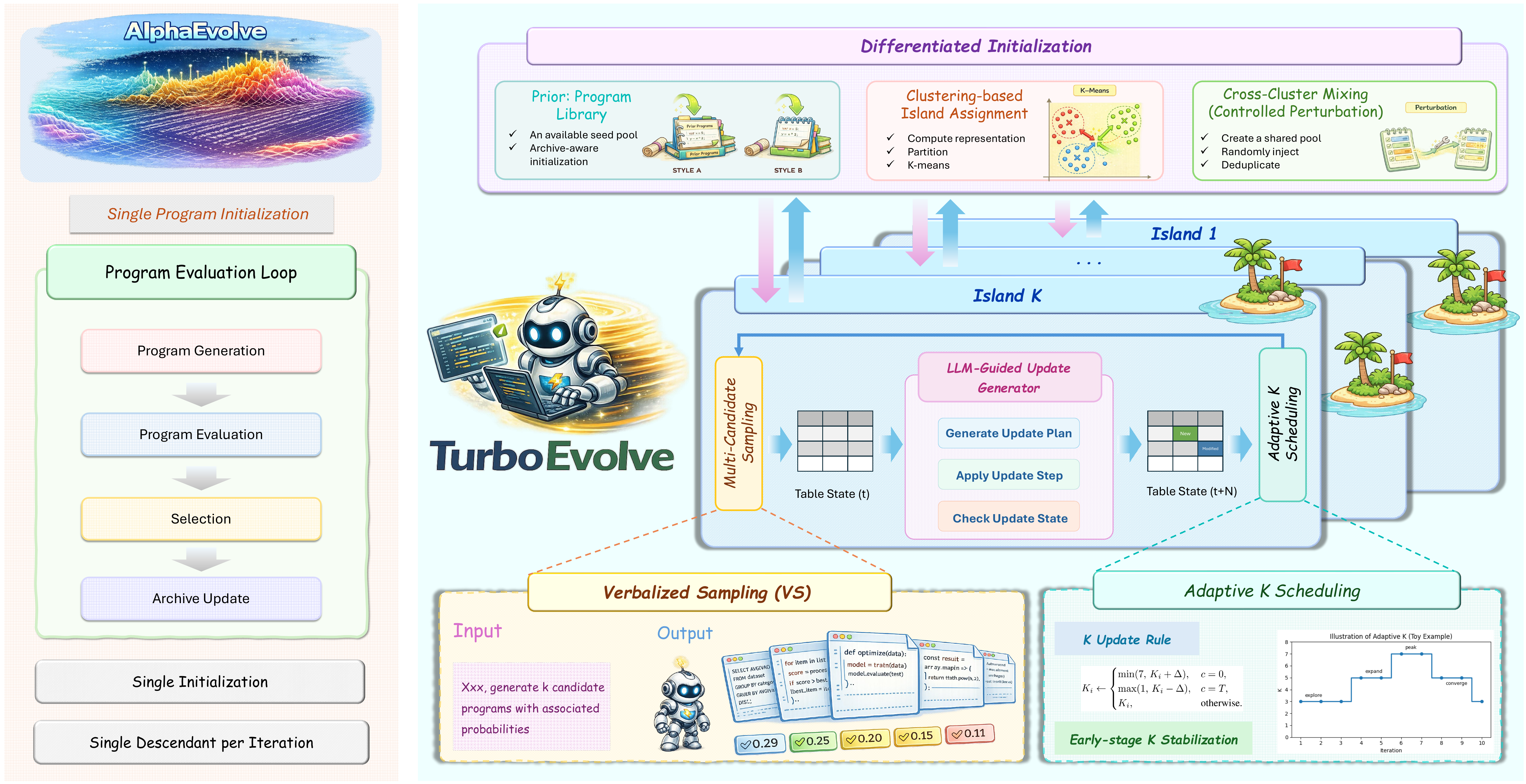}
    \caption{Comparison between \sysname{AlphaEvolve} and \sysname{TurboEvolve}. 
    \sysname{AlphaEvolve} uses single-program initialization and one descendant per iteration, whereas \sysname{TurboEvolve} adds island-aware initialization, Verbalized Sampling for multi-candidate generation, and adaptive $K$ scheduling.
    }
    \label{fig:framework}
\end{figure}

Built on \sysname{OpenEvolve}'s multi-island, archive-based program-evolution backbone~\citep{openevolve}, 
\sysname{TurboEvolve} extends the standard LLM generate--evaluate loop with three budget-critical yet often overlooked system-level controls: 
\textbf{Verbalized Sampling} for improving per-call candidate yield, 
\textbf{island-wise adaptive $K$ scheduling} for online island-level search-width allocation, 
and \textbf{island-aware initialization} for warm-start diversity from external solution pools (Fig.~\ref{fig:framework}).

\subsection{Generating \texorpdfstring{$K$}{K} Candidates via Verbalized Sampling}
\label{subsec:joint_generation}

\paragraph{Single-candidate bottleneck.}
Generating only one rewrite per evolutionary event makes LLM-driven program evolution sample-inefficient: a single invalid, redundant, or low-quality candidate can waste the prompt context, model call, and evaluation attempt while keeping the archive trajectory narrow. Repeating single-candidate calls under similar parent/inspiration contexts can also return near-duplicate rewrites, since each call is unaware of the alternatives generated in other calls. We therefore broaden each event into a joint set of alternatives, allowing the prompt to explicitly encourage distinct candidates within the same event.

\paragraph{VS-based multi-candidate generation.}
We instantiate this multi-offspring step with Verbalized Sampling (VS), asking the model to return $K$ distinct candidate programs in a single response. This design provides two benefits. 
\emph{(i) Candidate coverage.} Even if some returned programs are invalid, redundant, or low-quality, another candidate may still yield a useful archive update. We do not assume these candidates are independent; instead, the shared context and explicit distinctness requirement encourage complementary rewrites under the same parent/inspiration context. 
\emph{(ii) Prompt amortization.} Let $c_{\mathrm{ctx}}$ be the shared prompt cost, $c_{\mathrm{out}}$ the average output cost of one candidate, and $c_{\mathrm{out}}^{(K)}$ the output cost of returning $K$ candidates jointly. Then $K$ same-context single-candidate calls cost $C_{\mathrm{rep}}^{(K)}=K(c_{\mathrm{ctx}}+c_{\mathrm{out}})$, whereas one VS call costs $C_{\mathrm{VS}}^{(K)}=c_{\mathrm{ctx}}+c_{\mathrm{out}}^{(K)}$. Thus, VS is cheaper on the generation side whenever
\[
C_{\mathrm{VS}}^{(K)} < C_{\mathrm{rep}}^{(K)}
\quad \Longleftrightarrow \quad
c_{\mathrm{out}}^{(K)} < Kc_{\mathrm{out}} + (K-1)c_{\mathrm{ctx}} .
\]
This comparison concerns prompt-side amortization under a fixed evolutionary context, rather than sequential calls with intermediate feedback.

\paragraph{Evaluation and archive insertion.}

VS modifies only the proposal stage: returned programs are filtered, evaluated, and inserted into the island/archive map using the backbone's standard feasibility checks and update rules~\citep{openevolve}. The verbalized probabilities serve only as elicitation signals, not downstream selection scores.\footnote{We keep the first $K$ returned candidates. If fewer than $K$ candidates are returned, we accept them as-is; if the number falls below a small minimum threshold, we re-query for that round.} Prompt details are in Appendix~\ref{appendix:prompt_diff_circle_packing}.

\subsection{Adaptive Scheduling of \texorpdfstring{$K$}{K} at the Island Level}
\label{subsec:adaptive_k}
\paragraph{Static candidate-count limitation.}

Generating more candidates per call can improve exploration, but a larger $K$ is not uniformly beneficial under fixed budgets. Each additional candidate incurs generation and evaluation cost: a uniformly large $K$ can spend extra budget on redundant variants when search is already progressing, whereas a uniformly small $K$ can under-explore when an island stalls. Thus, $K$ should be an island-level search-width variable responsive to local progress rather than a fixed prompting hyperparameter.

\paragraph{Budget-aware search-width control.}
This perspective is inspired by dynamic parameter control and self-adjusting evolutionary algorithms,
where search parameters are adjusted using online feedback rather than fixed throughout optimization~\citep{karafotias2014parameter,doerr2019theory,rajabi2020self}. 
For a generic search state $s$, the budget--utility trade-off can be written as
\begin{equation}
K^\star(s)=\arg\max_{K\in\mathcal{K}} \left[r_s(K)-\lambda c(K)\right],
\label{eq:budget_obj}
\end{equation}
where $r_s(K)$ is the expected search utility of allocating $K$ candidates, $c(K)$ is the corresponding generation/evaluation cost, and $\lambda>0$ is a budget weight. 
Eq.~\eqref{eq:budget_obj} highlights that increasing $K$ is useful only when the additional search benefit justifies the additional cost. 
Operationally, we instantiate this trade-off with a lightweight controller: following the diminishing-return intuition detailed in Appendix~\ref{appendix:adaptive_k}, recent archive updates serve as a proxy for each island's local search regime.

\paragraph{Island-local controller.}
To operationalize this proxy, each island $i$ maintains its own adaptive $K_i$ over an
island-local window of $T$ proposal events. A proposal event includes any retries needed to
obtain its candidate set and contributes only one scheduler observation; all retry calls and
tokens are still included in cost and invocation counts.
Let $u_{i,\tau}\in\{0,1\}$ indicate whether event $\tau$ triggers any archive-cell replacement,
and define $c_i=\sum_{\tau=1}^{T}u_{i,\tau}$.
At the end of each window, the aggregate count $c_i$ determines the update:
\begin{equation}
K_i \leftarrow
\begin{cases}
\min(K_{\max},\,K_i+\Delta), & c_i=0,\\
\max(K_{\min},\,K_i-\Delta), & c_i=T,\\
K_i, & \text{otherwise.}
\end{cases}
\label{eq:update_rule}
\end{equation}
This implements the regime-dependent intuition above in a conservative way:
no replacement in a window increases $K_i$ for exploration; replacements in every event decrease
$K_i$ to reduce redundancy; otherwise $K_i$ is unchanged to avoid oscillation.
We start from $K_{\mathrm{init}}=5$ rather than $K_{\min}$ to provide early-stage search breadth.
Unless otherwise stated, $\mathcal{K}=\{1,3,5,7\}$ and $\Delta=2$, $T=3$.
The controller adapts search breadth; parent/inspiration selection determines the search direction.
Sensitivity to the candidate-count range and update window is reported in
Section~\ref{sec:adaptive_k_ablation}.

\subsection{Island-Aware Initialization from Seed Pools}

\providecommand{\Top}{\operatorname{Top}}
\providecommand{\Tail}{\operatorname{Tail}}
\providecommand{\Sample}{\operatorname{Sample}}
\providecommand{\ArchiveInsert}{\operatorname{ArchiveInsert}}
\newcommand{\algline}[2]{%
    \State \ensuremath{#1}\Comment{#2}%
}

\begin{algorithm}[!tbp]
\caption{Island-Aware Seed-Pool Initialization}
\label{alg:init}
\begin{algorithmic}[1]
\Require Scored seed pool $\mathcal{S}_{\mathrm{seed}}$, number of islands $M$, mixing ratio $d$, elite-protection ratio $\rho$
\Ensure Initialized island archives $\{\mathcal{A}_i\}_{i=1}^{M}$

\algline{\mathbf{Z}\leftarrow \mathrm{UniXcoder}(\mathcal{S}_{\mathrm{seed}})}
        {embed programs}
\algline{\{\mathcal{C}_i\}_{i=1}^{M}\leftarrow \mathrm{kMeans}(\mathbf{Z},M)}
        {split into modes}
\algline{\mathcal{P}\leftarrow \bigcup_{i=1}^{M}\Top_{\lceil d|\mathcal{C}_i|\rceil}(\mathcal{C}_i)}
        {shared elite pool}

\For{$i=1$ to $M$}
    \algline{n_i\leftarrow |\mathcal{C}_i|,\quad e_i\leftarrow \lceil \rho n_i\rceil}
            {size and protected elites}
    \algline{r_i\leftarrow \min\{\lfloor d n_i\rfloor,\ |\mathcal{P}\setminus\mathcal{C}_i|,\ n_i-e_i\}}
            {safe mixing count}
    \algline{\mathcal{E}_i\leftarrow \Top_{e_i}(\mathcal{C}_i)}
            {elite subset}
    \algline{\mathcal{R}_i\leftarrow \Sample_{r_i}(\mathcal{P}\setminus\mathcal{C}_i)}
            {cross-cluster imports}
    \algline{\mathcal{D}_i\leftarrow \Tail_{r_i}(\mathcal{C}_i\setminus\mathcal{E}_i)}
            {replaceable tail}
    \algline{\widetilde{\mathcal{C}}_i\leftarrow (\mathcal{C}_i\setminus\mathcal{D}_i)\cup \mathcal{R}_i}
            {mixed seed set}
    \algline{\mathcal{A}_i\leftarrow \ArchiveInsert(\widetilde{\mathcal{C}}_i)}
            {archive initialization}
\EndFor

\State \Return $\{\mathcal{A}_i\}_{i=1}^{M}$
\end{algorithmic}
\end{algorithm}

\paragraph{Warm-start allocation problem.}

Many program-evolution workloads provide a scored pool of candidate programs from prior runs. In a multi-island setting, however, naively reusing this pool can waste budget: identical starts encourage redundant local search, while random mixing can blur distinct solution modes. We therefore seek \emph{mode-separated yet partially shared} warm starts: each island begins from a distinct seed mode, while limited cross-cluster sharing transfers useful motifs. Given a seed pool $\mathcal{S}_{\mathrm{seed}}$ and $M$ islands, Algorithm~\ref{alg:init} initializes archives balancing mode separation, elite preservation, and controlled sharing.

\paragraph{Clustered allocation and elite-preserving mixing.}

Algorithm~\ref{alg:init} first embeds seed programs and clusters them into $M$ mode-specific groups, assigning one group to each island. To avoid over-isolating islands, it builds a shared elite pool and injects a small number of cross-cluster seeds by replacing only low-ranked non-elites, while protecting the top $\lceil \rho n_i\rceil$ local seeds. Here, $\Top$ and $\Tail$ rank programs by task score, and $\ArchiveInsert$ denotes the same feasibility checks and map/grid insertion policy used during evolution. The default settings for $d$ and $\rho$ are $d=0.3$ and $\rho=0.1$.

\section{Experimental Setup}
\label{sec:experimental_setup}

Our experiments are organized around six questions:
\par\smallskip
\begingroup
\setlength{\tabcolsep}{0pt}
\renewcommand{\arraystretch}{1.05}
\noindent
\begin{tabularx}{\textwidth}{@{}p{0.052\textwidth}@{\hspace{0.45em}}X@{}}

\textbf{RQ1} & \emph{Cost-matched performance and consistency.}
Under matched API-cost budgets, does \sysname{TurboEvolve} obtain stronger best-so-far results and more consistent performance across repeated runs than controlled baselines? \\[2pt]

\textbf{RQ2} & \emph{Backbone-paired efficiency.}
At the same evaluated-program horizon, how much API cost and how many LLM invocations does \sysname{TurboEvolve} save relative to \sysname{OpenEvolve}? \\[2pt]

\textbf{RQ3} & \emph{Joint versus separate calls.}
Does generating candidates in one call outperform separate single-candidate calls with the same context under matched token budgets or nearly equal evaluation counts? \\[2pt]

\textbf{RQ4} & \emph{Candidate-set utility.}
Do candidates from one call provide distinct and useful search alternatives? \\[2pt]

\textbf{RQ5} & \emph{Candidate-count scheduling.}
Does adapting the number of candidates per call improve the cost--quality trajectory over a fixed count, and how sensitive is performance to the range and update window? \\[2pt]

\textbf{RQ6} & \emph{Solution-pool initialization.}
When a reusable seed pool is available, does island-aware allocation improve warm-start initialization?
\end{tabularx}
\par
\endgroup

\subsection{Tasks}

We evaluate eight executable program-optimization tasks. Five mathematical/algorithmic discovery
tasks come from \sysname{AlphaEvolve}~\citep{novikov2025alphaevolve}: first and second autocorrelation
inequalities (\taskname{Autocorr\_First} and \taskname{Autocorr\_Second}), the uncertainty inequality (\taskname{Uncertainty}),
packing 26 circles in a unit square (\taskname{CirclePack\_26}), and Erd\H{o}s minimum overlap (\taskname{Erdos\_Overlap}).
Three KernelBench Level 2 tasks~\citep{ouyang2025kernelbench} optimize GPU operator sequences:
\taskname{Matmul\_SwishGN}, \taskname{Conv3d\_Mish}, and
\taskname{Conv2d\_BN}.
GN and BN denote group normalization and batch normalization, respectively.
All evaluators check feasibility or
correctness and return an objective to minimize or maximize. Exact task identifiers, objectives,
and evaluator details are given in Appendix~\ref{appendix:task}.

\subsection{Protocol: Models, Budgets, and Reporting}
\label{subsec:protocol}

\textbf{Models and alignment.} Each main method--task--backend configuration is run five times
with Gemini-2.5-Flash and Gemini-3-Flash (\texttt{gemini-3-flash-preview}). Across backends we keep
prompt templates, method hyperparameters, evaluators, and stopping rules fixed. Within each block,
methods share the evaluator and budget protocol; differences concern their algorithmic design.
The main comparisons use cold starts; warm-start initialization is studied separately.

\looseness=1
\textbf{Budgets.} For each task and Gemini backend, \sysname{TurboEvolve} is run to 800 evaluated programs for
mathematical tasks or 100 for KernelBench. Its mean API cost over five runs sets the budget for each
baseline, which stops on reaching that limit. Costs include logged input/output tokens, available
provider-side prompt/prefix caching, and retries, under the same accounting rule for every method
within a backend block. A complementary \sysname{OpenEvolve} diagnostic uses the same program horizon
instead of the same API-cost budget, measuring the calls and cost needed to evaluate that many
programs. API cost excludes local evaluator compute.
Supplementary controls separately match input tokens, output tokens, cached-input usage,
cache-aware API cost, evaluator compute, and \mbox{active wall time (Appendix~\ref{appendix:resources})}.

\input{tabs/main_result.tex}

\textbf{Reporting.} The main table reports the best score over five runs; robustness plots report
the mean and standard deviation of final per-run best scores. Appendix~\ref{appendix:budget_protocol}
details budget overshoot, failed-evaluation counting, stopping rules, and implementation settings.
Supplementary controls use the reporting conventions specified with their results.

\subsection{Comparisons, Ablations, and Initialization}
\label{subsec:ablations}

\textbf{Comparisons.} We rerun \sysname{OpenEvolve}~\citep{openevolve}, \sysname{LoongFlow}~\citep{wan2025loongflow},
and \sysname{PartEvo}~\citep{hu2025partition}\footnote{
\sysname{PartEvo}'s original evaluation reports consistent improvements over
\sysname{FunSearch}~\citep{romera2024mathematical}, \sysname{EoH}~\citep{liu2024evolution}, and \sysname{ReEvo}~\citep{ye2024reevo}.
} under the matched-cost protocol. \sysname{OpenEvolve} also provides the
single-descendant backbone for the equal-horizon diagnostic. Rankings compare controlled reruns within each backend. \sysname{AlphaEvolve}~\citep{novikov2025alphaevolve} is a reported reference;
no public implementation supports matched reruns.

\textbf{Independent-$K$ control.} On \taskname{Uncertainty} using Gemini-3-Flash, Independent-$K$
replaces each \sysname{TurboEvolve} joint call with $K$ separate single-candidate calls. Within each group,
the parent, inspirations, and archive are frozen, with no intermediate evaluator feedback.
This differs from ordinary sequential \sysname{OpenEvolve}, which may update its context after each candidate.
We compare separately matched input-, output-, and total-token budgets and nearly equal
program counts (Section~\ref{sec:joint_control}). For the latter control, we truncate each run
at its last complete outer iteration with at most 800 evaluated programs, avoiding partial
candidate groups.

\textbf{Candidate-set diagnostics.} On \taskname{CirclePack\_26}, prefix replay retains the first $m\leq7$
candidates in returned order from each fixed $K=7$ generated set. We examine code differences
via normalized Levenshtein distance and candidate outcomes by returned position
(Section~\ref{sec:candidate_utility}; Appendix~\ref{appendix:more_analysis}).

\textbf{Adaptive $K$ ablation.} On \taskname{CirclePack\_26} with Gemini-2.5-Flash, we compare adaptive
$K$ with fixed $K\in\{3,5,7\}$. All variants share the evaluator, VS output format, and 800-program
horizon, isolating state-dependent candidate-count control.
Mean@800 denotes the across-run mean of the best score at the 800-program horizon;
trajectory AUC denotes area under the curve.
Sensitivity tests on \taskname{Erdos\_Overlap} and \taskname{CirclePack\_26} separately increase
$K_{\max}$ from 7 to 9 and $T$ from 3 to 5 (Section~\ref{sec:adaptive_k_ablation}).

\textbf{Initialization.} A seed pool is collected automatically from separate cold-start runs, with
no target-run trajectory reuse or hand-picked seeds. Every allocation strategy receives the same
fixed pool. With the remaining pipeline fixed, we compare random allocation (random),
$k$-means (kmeans), and $k$-means with elite-preserving cross-cluster mixing (kmeans+elite),
using either the full pool or the bottom 80\%
after removing the highest-scoring 20\% of seeds.

\section{Results}
\label{sec:results}

\subsection{Cost-Matched Performance and Repeated-Run Consistency}
\label{sec:results_sota}

Table~\ref{tab:main_results} reports the best score obtained across five runs for each controlled
method under the matched API-cost protocol. \sysname{TurboEvolve} \textbf{ranks first on 15 of 16
controlled task--backend pairs} and second on the remaining one. This overall advantage is observed
with both Gemini-2.5-Flash and Gemini-3-Flash. It also spans mathematical discovery and GPU-kernel
optimization, covering the different validity constraints and optimization objectives of the two task
families. The gains are especially pronounced on the GPU-kernel optimization tasks.

\begin{figure}[t]
    \centering
    \includegraphics[width=\textwidth]{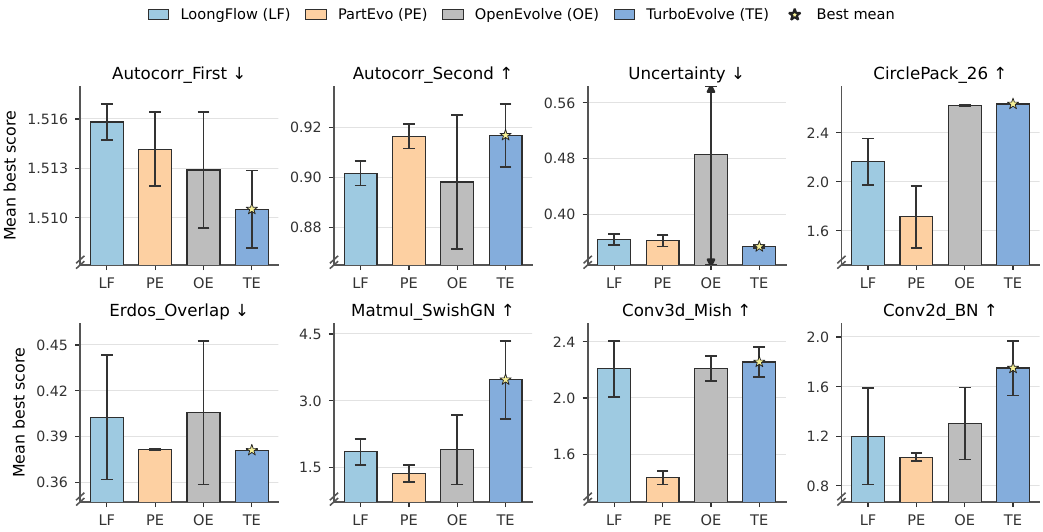}
    
    \caption{Run-to-run consistency under Gemini-2.5-Flash. Bars show the mean of final per-run best scores over five runs, and error bars show standard deviation.}
    \label{fig:robustness_g25}
\end{figure}

\begin{samepage}
Performance across repeated runs is important when only a few search restarts are affordable.
Figure~\ref{fig:robustness_g25} reports the results of \sysname{TurboEvolve} and the baselines on all
eight tasks using Gemini-2.5-Flash over five runs. Under the same API-cost budget,
\sysname{TurboEvolve} \textbf{achieves the best average performance on all eight tasks} and has
smaller standard deviations than \sysname{OpenEvolve} on most tasks. These results show that
\sysname{TurboEvolve} can find better solutions more consistently within a fixed per-run allowance.
The same search budget delivers stronger average outcomes, with less run-to-run variation than
\sysname{OpenEvolve} on most tasks. The corresponding results with Gemini-3-Flash are provided in
Appendix~\ref{appendix:robustness_g3}.
\par
\end{samepage}

\subsection{Backbone-Paired Efficiency at Equal Evaluated-Program Horizons}
\label{sec:backbone_efficiency}

We compare \sysname{TurboEvolve} and \sysname{OpenEvolve} at the same evaluated-program horizon to
assess LLM calls, API cost, and endpoint quality. This paired diagnostic uses the same backbone,
evaluator, and archive-update protocol.

\begin{figure}[t]
\centering
\sbox{\efficiencysummarybox}{%
\begin{minipage}{0.63\textwidth}
\centering
\input{tabs/efficiency_full.tex}\par
\end{minipage}}
\begin{minipage}[t]{0.63\textwidth}
\vspace{0pt}
\centering
\textbf{(A) Endpoint summary}\par\smallskip
\usebox{\efficiencysummarybox}
\end{minipage}\hfill
\begin{minipage}[t]{0.35\textwidth}
\vspace{0pt}
\centering
\textbf{(B) Representative trajectories}\par\smallskip
\parbox[c][\dimexpr\ht\efficiencysummarybox+\dp\efficiencysummarybox\relax][c]{\linewidth}{%
\centering
\includegraphics[width=\linewidth]{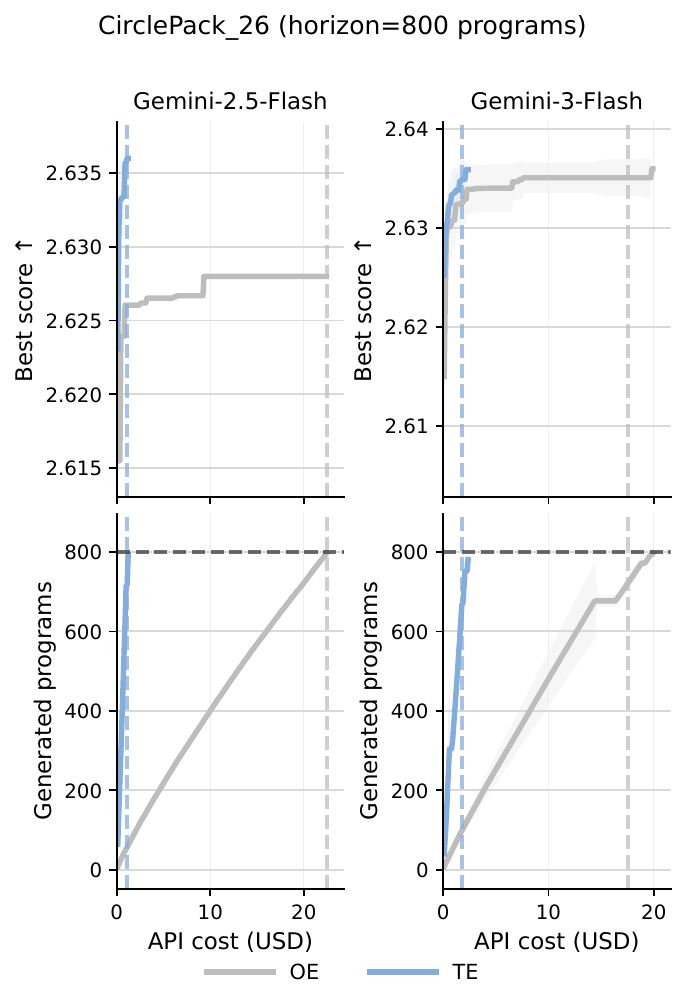}}
\end{minipage}
\caption{Backbone-paired efficiency at equal evaluated-program horizons: 800 programs for
mathematical tasks and 100 for KernelBench tasks. (A) Endpoint score, LLM calls, and API cost.
OE: \sysname{OpenEvolve}; TE: \sysname{TurboEvolve}. Bold marks the better score or lower resource use within each pair.
(B) Representative \taskname{CirclePack\_26} trajectories. Columns: Gemini-2.5-Flash and Gemini-3-Flash;
rows: best score and generated programs versus API cost.}
\label{fig:backbone_efficiency}
\end{figure}

Figure~\ref{fig:backbone_efficiency}(A) shows that, averaged equally over 16 task--backend pairs,
\sysname{TurboEvolve} \textbf{reduces LLM calls by 73.3\% and API cost by 71.3\%}, while improving
endpoint scores on 13 of 16 pairs. It completes the same number of program evaluations with
substantially fewer model calls and lower API expenditure. These reductions accompany stronger
endpoint solutions in most settings, combining efficient candidate generation with effective search.
Figure~\ref{fig:backbone_efficiency}(B) illustrates how this advantage develops during the search
on \taskname{CirclePack\_26}. Under both backends, \sysname{TurboEvolve} reaches the target evaluated-program
horizon at lower API cost. Improvements in best-so-far quality also appear at lower accumulated
expenditure, showing progress toward stronger solutions with less spending along the trajectory.
Additional task trajectories are provided in Appendix~\ref{appendix:efficiency_trajectories}.

\looseness=1
The API-cost reductions include the prompt/prefix caching benefits available to every method
under the provider's billing policy. A separate shared-context analysis estimates 60.1--85.2\%
lower repeated input-token usage (median 76.5\%) than generating the retained candidates in separate
calls. This reduction comes from supplying shared context once per candidate set.
To distinguish lower generation cost from improvements in search quality,
Section~\ref{sec:joint_control} compares outcomes under matched token budgets and at nearly equal
program counts, while Section~\ref{sec:candidate_utility} examines the utility of additional
candidates within a jointly generated set. Appendix~\ref{appendix:resources} provides the accounting
details and comparisons under \mbox{alternative resource constraints}.

\subsection{TurboEvolve versus Independent-\texorpdfstring{$K$}{K}}
\label{sec:joint_control}

The efficiency results motivate a further question: does \sysname{TurboEvolve}'s joint generation provide useful candidate
sets beyond amortizing repeated prompt context? We compare \sysname{TurboEvolve} with an Independent-$K$ control
on \taskname{Uncertainty} using Gemini-3-Flash, following the protocol in Section~\ref{subsec:ablations}.

Under separately matched input-, output-, and total-token budgets, \sysname{TurboEvolve} evaluates
approximately $3.9$--$4.7\times$ as many programs and obtains a 5.8--6.6\% lower mean objective than
Independent-$K$. These controls establish stronger performance per token but alone cannot separate
candidate quality from the benefit of evaluating more programs. Input-token matching emphasizes
shared-context costs in separate calls; output-token matching constrains generated content.
We therefore also compare nearly equal program counts; token-budget details appear in
Appendix~\ref{appendix:independent_k}.

\begin{table}[t]
\noindent
\begin{minipage}[t]{\textwidth}
\vspace{0pt}
\noindent
\begin{minipage}[t]{0.55\textwidth}
\vspace{0pt}
\centering
\settowidth{\controlcaptionwidth}{\taskname{Uncertainty} (Gemini-3-Flash), with near-equal}
\addtolength{\controlcaptionwidth}{1pt}
\begin{minipage}[t]{\controlcaptionwidth}
\vspace{0pt}
\captionof{table}{\raggedright\sysname{TurboEvolve} vs.\ Independent-$K$ on\newline
\taskname{Uncertainty} (Gemini-3-Flash), with near-equal\newline
program counts; mean $\pm$ standard deviation.}
\label{tab:independent_horizon}
\end{minipage}\par
\end{minipage}\hfill
\begin{minipage}[t]{0.42\textwidth}
\vspace{0pt}
\captionof{table}{Prefix coverage on \taskname{CirclePack\_26}: percentage of 495 $K=7$ calls
containing an improving or valid program.}
\label{tab:prefix_utility}
\end{minipage}
\par\vspace{\baselineskip}
\nointerlineskip
\noindent
\begin{minipage}[t]{0.55\textwidth}
\vspace{0pt}
\centering
\setlength{\tabcolsep}{6pt}
\begin{tabular}{@{}lcc@{}}
\toprule
Method & \makecell{Evaluated\\programs} & Objective $\downarrow$\\
\midrule
\sysname{TurboEvolve} & $797.4\pm2.4$ & $\mathbf{0.330551\pm0.002858}$\\
\addlinespace[3pt]
Independent-$K$ & $798.3\pm0.5$ & $0.347405\pm0.004285$\\
\bottomrule
\end{tabular}
\end{minipage}\hfill
\begin{minipage}[t]{0.42\textwidth}
\vspace{0pt}
\centering
\setlength{\tabcolsep}{4pt}
\begin{tabular}{@{}lrr@{}}
\toprule
Coverage\vphantom{\makecell{Evaluated\\programs}} & First & All seven\\
\midrule
Improving program & 9.1\% & 32.7\%\\
\addlinespace[3pt]
Valid program & 14.5\% & 91.1\%\\
\bottomrule
\end{tabular}
\end{minipage}
\end{minipage}
\end{table}

Under the complete-iteration truncation protocol in Section~\ref{subsec:ablations},
Table~\ref{tab:independent_horizon} shows almost identical horizons, yet
\sysname{TurboEvolve} achieves a \textbf{4.85\% lower mean objective}. Its slightly lower evaluation count
marginally favors Independent-$K$. This lower objective on \taskname{Uncertainty} with Gemini-3-Flash
shows that the quality advantage cannot be explained solely by evaluating more programs.

\subsection{Candidate-Set Utility and Search Dynamics}
\label{sec:candidate_utility}

To examine the value of additional candidates on \taskname{CirclePack\_26}, we use the $K=7$
prefix replay described in Section~\ref{subsec:ablations}.
Table~\ref{tab:prefix_utility} summarizes first-candidate and full-set coverage.
Figure~\ref{fig:analysis_1} reports the resulting improvement coverage and best
score change, stratified by inspiration tier; Appendix~\ref{sec:more_counterfactual} gives the protocol.

\par\medskip
\noindent\begin{minipage}{\textwidth}
\centering
\includegraphics[width=0.95\linewidth]{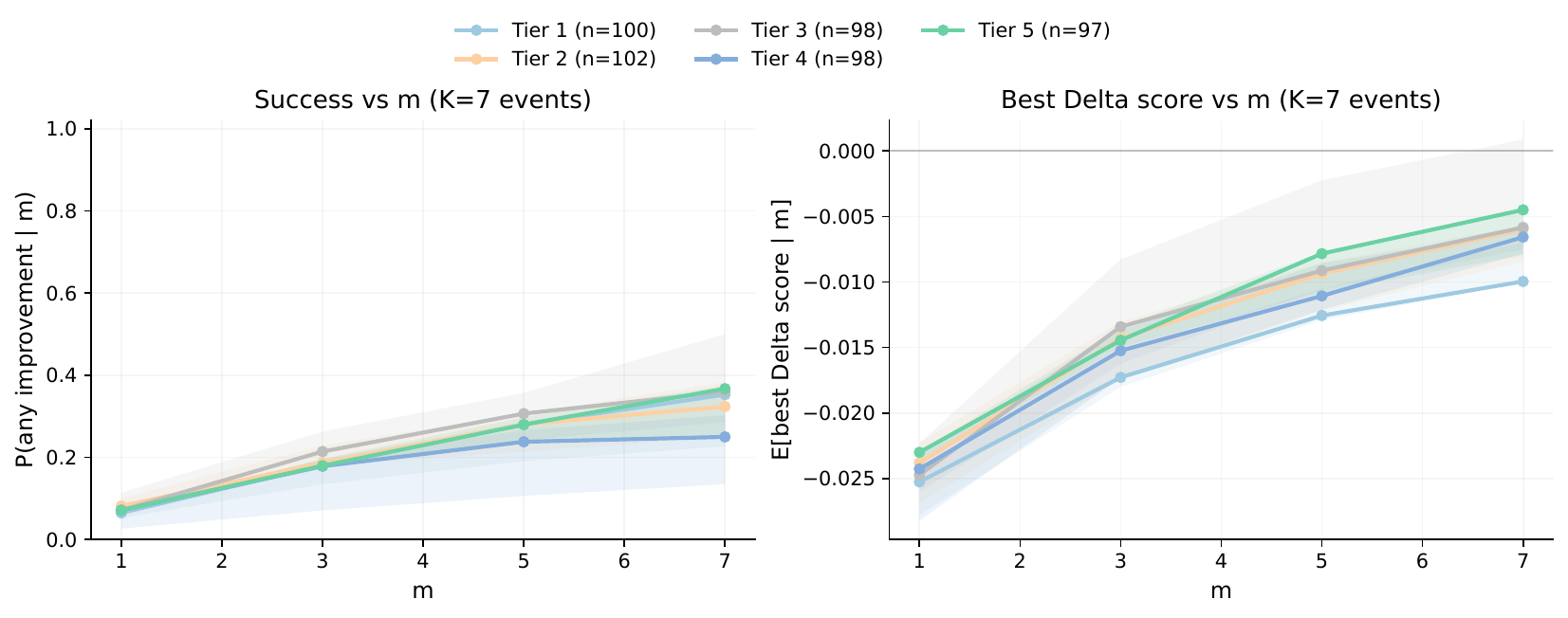}
\captionof{figure}{Within-event prefix replay on \taskname{CirclePack\_26} ($K=7$).
Left: probability that a retained prefix contains a valid improvement.
Right: best score change $\max_{j\leq m}\Delta_j$ relative to the parent; negative values indicate regression.
Curves group events by inspiration-score quintile; lines show medians across runs and bands show IQR.}
\label{fig:analysis_1}
\end{minipage}
\par\medskip

Candidates generated from a shared context may be correlated, so we examine syntactic differences
separately from search utility. Across 1,023 multi-candidate calls, 2,299 recovered within-call
pairs contain no exact duplicates and 1.91\% near-duplicates (normalized Levenshtein distance
$\leq0.10$). Moreover, 96.7\% of calls contain no near-duplicate pair, and the mean within-call
normalized distance is 0.391.

Table~\ref{tab:prefix_utility} shows $3.6\times$ higher improvement coverage, while larger marginal gains
with higher-scoring inspirations (Figure~\ref{fig:analysis_1}) suggest that later candidates' benefits depend on prompt context.
Coverage and prefix maxima are nondecreasing by construction; observed gains quantify within-call utility,
not the causal effect of larger $K$.

The rank profile (Appendix~\ref{sec:more_rank_profile}) shows complementary roles in $K=7$ calls on
\taskname{CirclePack\_26}: head candidates (1--2) yield the strongest immediate gains when valid;
tail candidates (6--7) are more often valid and farther from the primary parent in archive space,
but yield weaker valid-only gains. Thus, returned order is not monotonic in quality, nor does greater
archive distance alone reliably predict larger immediate gains.

Code distances imply neither semantic nor statistical independence; these tests do not isolate
probability elicitation, used only for generation (Section~\ref{subsec:joint_generation}). Scheduler theory requires
explicit utility assumptions; warm-start tests control seed access to our cold-start pools,
leaving generalization to external pools open.

\newlength{\tablefouroriginaltextfloatsep}
\setlength{\tablefouroriginaltextfloatsep}{\textfloatsep}
\textfloatsep=12pt plus 2pt\relax
\AddToHookNext{shipout/after}{\global\textfloatsep=\tablefouroriginaltextfloatsep\relax}
\begin{table}[!b]
\setlength{\abovecaptionskip}{0pt}
\centering
\caption{Sensitivity to scheduler range and update window. Entries are mean $\pm$
standard deviation; arrows indicate the optimization direction.}
\label{tab:scheduler_sensitivity}
\par\vspace{\baselineskip}
\setlength{\tabcolsep}{4pt}
\begin{tabular*}{\linewidth}{@{\extracolsep{\fill}}lcccc@{}}
\toprule
Variant & $T$ & Candidate counts $K$ &
\taskname{Erdos\_Overlap}$\downarrow$ &
\taskname{CirclePack\_26}$\uparrow$\\
\midrule
Default & 3 & $\{1,3,5,7\}$ & $0.380927\pm0.000041$ & $2.635114\pm0.001690$\\
Wider range & 3 & $\{1,3,5,7,9\}$ & $0.380985\pm0.000069$ & $2.634228\pm0.002474$\\
Slower update & 5 & $\{1,3,5,7\}$ & $0.380944\pm0.000042$ & $2.632477\pm0.002243$\\
\bottomrule
\end{tabular*}
\par
\par\vspace{\baselineskip}
\end{table}

\subsection{Ablation on Adaptive \texorpdfstring{$K$}{K} Scheduling}
\label{sec:adaptive_k_ablation}

\noindent
\begin{minipage}[t]{0.465\textwidth}
\vspace{0pt}
Following the protocol in Section~\ref{subsec:ablations}, we compare adaptive $K$ with fixed
$K\in\{3,5,7\}$ on \taskname{CirclePack\_26}. This ablation asks whether state-dependent
candidate-count control provides a better search trajectory than using one fixed candidate count
throughout the run.
\par\smallskip

Figure~\ref{fig:adaptive_k_ablation} shows that adaptive $K$ matches the best final Mean@800 and
improves search efficiency, with the \textbf{highest AUC and earliest attainment of all tested
quality thresholds}. For example, reaching 2.635 requires only 212.3 generated
programs with adaptive scheduling, compared with 369.0, 560.0, and 567.0 programs for fixed
$K=3,5,7$, respectively. Even against the best fixed final-quality setting ($K=3$), adaptive $K$
reaches 2.635 with 42.5\% fewer generated programs. This supports the intended role of adaptive
scheduling: it is less a final-score booster than a \textbf{budget-allocation mechanism} that preserves
final quality while reducing the need to tune a single fixed candidate count for the entire run.
\end{minipage}\hfill
\begin{minipage}[t]{0.515\textwidth}
\vspace{0pt}
\centering
\textbf{(A) Search trajectory}
\par
\includegraphics[width=\linewidth]{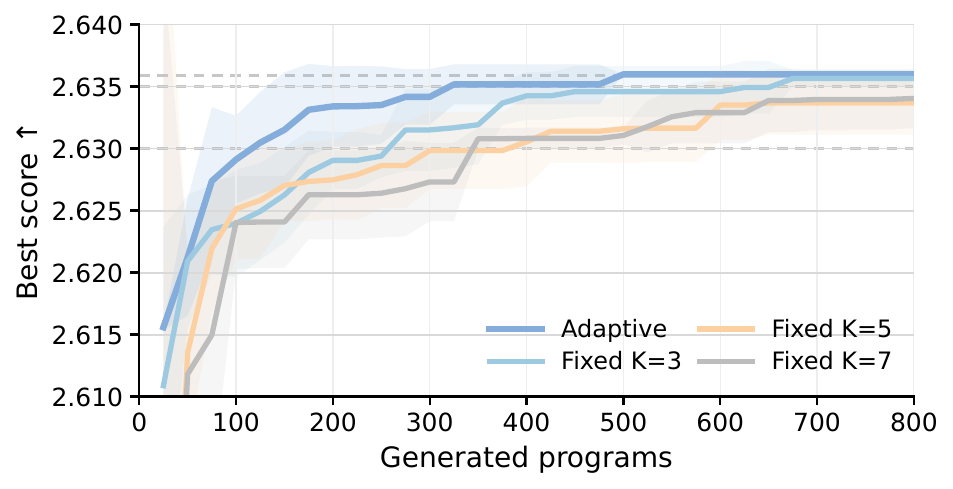}
\par\smallskip
\textbf{(B) Summary statistics}\par\smallskip
\setlength{\tabcolsep}{2.4pt}
\begin{tabular}{@{}lrrrr@{}}
\toprule
Metric & $K=3$ & $K=5$ & $K=7$ & Adaptive\\
\midrule
Mean@800 $\uparrow$ & 2.635983 & 2.635756 & 2.635981 & \textbf{2.635983}\\
AUC $\uparrow$ & 2.631538 & 2.628282 & 2.626791 & \textbf{2.633694}\\
Reach 2.63 $\downarrow$ & 229.0 & 290.0 & 357.0 & \textbf{107.3}\\
Reach 2.635 $\downarrow$ & 369.0 & 560.0 & 567.0 & \textbf{212.3}\\
Reach 2.6359 $\downarrow$ & 369.0 & 592.5 & 665.0 & \textbf{212.3}\\
\bottomrule
\end{tabular}
\captionof{figure}{Adaptive $K$ ablation with Gemini-2.5-Flash.
Thresholds count programs needed to reach each score.}
\label{fig:adaptive_k_ablation}
\end{minipage}
\par

\noindent
We also test a wider adaptive range (allowing but not requiring $K=9$ per call)
and a longer update window ($T$ from 3 to 5), on \taskname{Erdos\_Overlap} and
\taskname{CirclePack\_26} (Table~\ref{tab:scheduler_sensitivity}).
The default has the strongest mean objective on both tasks, with no gain from widening the range.
The objective varies modestly on \taskname{Erdos\_Overlap}, whereas slower adaptation
has a clearer effect on \taskname{CirclePack\_26}, lowering its mean from 2.635114 to 2.632477.
These checks support adapting breadth to recent feedback and show sensitivity to window length,
without proving controller optimality or replacing a broader hyperparameter study.
\par

\subsection{Initialization Study}
\label{sec:results_init}

\begin{figure}[htbp]
    \centering
    \includegraphics[width=\linewidth]{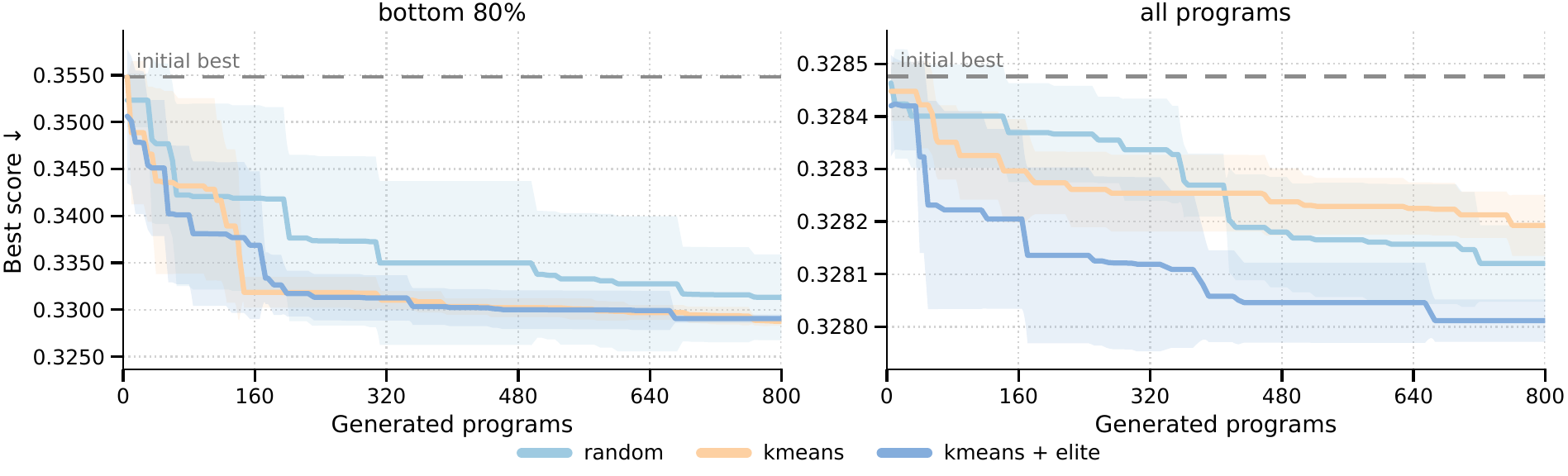}
    
    \caption{\textbf{Warm-start initialization on \taskname{Uncertainty}.}
    \sysname{TurboEvolve} with different seed-pool allocations: random, kmeans, and kmeans+elite.
    Left: bottom 80\% pool (top 20\% removed). Right: full pool.
    Curves show best-so-far objective across runs vs.\ evolved programs.}

    \label{fig:initialization}
\end{figure}

Figure~\ref{fig:initialization} shows that warm-start effects extend beyond short-horizon acceleration
and depend on pool quality and allocation. With the full pool, a strong initial incumbent limits headroom,
so all three strategies mainly refine solutions with only mild separation.

After removing the top 20\% seed programs, allocation becomes more consequential:
\textbf{structured allocation, especially kmeans+elite, sustains progress more reliably and
reaches a better final level} than random allocation. Since all strategies receive the same fixed
pool, this comparison concerns how seeds are distributed across islands, rather than access to
different initial solutions. The contrast between the full and reduced pools also makes clear that
the value of allocation depends on the quality and diversity of the available seeds.

Comparing kmeans with kmeans+elite links these results to allocation design: both use representation-based
clustering, while kmeans+elite also shares selected seeds across clusters and preserves local elites.
The results support distinct starting groups with selective sharing, retaining strong local seeds while
accessing useful alternatives from other groups.

This pattern suggests that the main value of island-aware initialization is controlled diversity:
rather than uniformly speeding up every phase, it improves the chance of entering productive
basins and maintaining improvement momentum after the warm start. Thus, seed-pool allocation acts
as a robustness lever for warm-start evolution, not merely as an initialization shortcut.
Responsible-use considerations for seed-pool initialization are discussed in
Appendix~\ref{appendix:seed_pool_responsible_use}.

\section{Conclusion}
\label{sec:conclusion}
We introduced \sysname{TurboEvolve}, a budget-aware framework for executable-evaluator-driven program evolution under costly LLM calls. Built on a multi-island backbone, \sysname{TurboEvolve} combines Verbalized Sampling for multi-candidate proposal, island-local adaptive scheduling of the candidate count $K$, and representation-based seed-pool allocation for mode-separated warm starts. We further curate a reusable solution-pool dataset to support warm-start initialization studies. Across mathematical discovery and GPU-kernel optimization tasks, \sysname{TurboEvolve} achieves stronger cost--quality trade-offs, lower model usage, and more stable performance under matched API-cost budgets. The current design keeps budget control lightweight, reacting to recent archive updates rather than explicitly modeling long-horizon search history, planning context, or recurring failure patterns. Future work can integrate planning- and memory-based evolution, calibrated within-call candidate signals, and stronger archive/diversity management for program search under controllable multi-resource budgets.

\subsubsection*{Acknowledgments}
This work was supported by the Guangdong Basic and Applied Basic Research Foundation (Grant No. 2026A1515011579), the HKUST-HKUST(GZ) 1+1+1 Joint Funding Program (Grant No. C\_2025\_031), and the Guangzhou-HKUST(GZ) Joint Funding Program (Grant No. 2023A03J0008), Education Bureau of Guangzhou Municipality. This work was also supported by Jiangsu Industrial Technology Research Institute (JITRI) and Wuxi National High-Tech District (WND).

\bibliographystyle{tmlr}
\bibliography{refs}

\newpage
\appendix

\section{Budget-Aware Interpretation of Adaptive \texorpdfstring{$K$}{K}}
\label{appendix:adaptive_k}

We expand the budget-aware intuition in Section \ref{subsec:adaptive_k} and interpret adaptive $K$ as a form of
\emph{state-dependent search-width control}. In LLM-driven program evolution, $K$ determines how
many candidate programs are generated within one LLM call, and therefore controls the per-round
branching width as well as the associated generation/evaluation cost. Since both model calls and
executable evaluations are expensive, choosing $K$ involves a utility--cost trade-off rather than a
monotonic ``larger is always better'' effect.

\paragraph{State-dependent candidate allocation.}
Let $s$ denote the current search regime, e.g., a \emph{progress regime} in which the current search
direction remains productive, or a \emph{stagnation regime} in which broader branching is more
valuable. For a given regime $s$, consider the budget-aware objective
\begin{equation}
K^\star(s)=\arg\max_{K\in\mathcal{K}} \left[r_s(K)-\lambda c(K)\right],
\label{eq:appendix_budget_obj}
\end{equation}
where $r_s(K)$ is the expected search utility of allocating $K$ candidates, $c(K)$ is the
corresponding generation/evaluation cost, and $\lambda>0$ is a budget weight.

To make the intuition explicit, we consider a continuous relaxation $K\in\mathbb{R}_+$.
Assume that the search utility is increasing with diminishing returns:
\begin{align}
\frac{\partial r_s}{\partial K}(K) &> 0,
\label{eq:mono_gain}\\
\frac{\partial^2 r_s}{\partial K^2}(K) &< 0,
\label{eq:diminishing_return}
\end{align}
and that the cost is increasing with nondecreasing marginal cost:
\begin{align}
c'(K) &> 0,
\label{eq:cost_increasing}\\
c''(K) &\ge 0.
\label{eq:cost_convex}
\end{align}
We also assume that the marginal utility of additional candidates is larger under stagnation than
under sustained progress:
\begin{equation}
\frac{\partial r_{\mathrm{stag}}}{\partial K}(K)
>
\frac{\partial r_{\mathrm{prog}}}{\partial K}(K),
\qquad \text{for relevant } K.
\label{eq:regime_gap}
\end{equation}
Intuitively, when the search stalls, additional branching has higher value because it can help escape
the current basin; when the search is already progressing steadily, the same extra candidates are more
likely to be redundant.

\paragraph{Ordering of regime-specific optimal $K$.}
Under the continuous relaxation, define
\[
F_s(K)=r_s(K)-\lambda c(K).
\]
By Eqs.~\eqref{eq:diminishing_return} and~\eqref{eq:cost_convex}, $F_s$ is concave. Assume each
regime admits a unique interior optimum. Then $K^\star(s)$ satisfies the first-order condition
\begin{equation}
\frac{\partial r_s}{\partial K}\bigl(K^\star(s)\bigr)
=
\lambda c'\bigl(K^\star(s)\bigr).
\label{eq:foc}
\end{equation}

\paragraph{Proposition 1.}
Under Eqs.~\eqref{eq:mono_gain}--\eqref{eq:regime_gap} and the unique-interior-optimum
assumption above, the regime-specific optimum under stagnation is no smaller than the optimum
under sustained progress:
\begin{equation}
K^\star(\mathrm{stag}) \ge K^\star(\mathrm{prog}).
\label{eq:ordering}
\end{equation}

\paragraph{Proof sketch.}
Let $K_p=K^\star(\mathrm{prog})$. Since $K_p$ is an interior optimum for the progress regime,
\[
F'_{\mathrm{prog}}(K_p)
=
\frac{\partial r_{\mathrm{prog}}}{\partial K}(K_p)-\lambda c'(K_p)
=0.
\]
Using Eq.~\eqref{eq:regime_gap}, we have
\[
F'_{\mathrm{stag}}(K_p)
=
\frac{\partial r_{\mathrm{stag}}}{\partial K}(K_p)-\lambda c'(K_p)
>
\frac{\partial r_{\mathrm{prog}}}{\partial K}(K_p)-\lambda c'(K_p)
=0.
\]
Because $F_{\mathrm{stag}}$ is concave, a positive derivative at $K_p$ implies that its maximizer
cannot lie to the left of $K_p$. Therefore,
$K^\star(\mathrm{stag})\ge K^\star(\mathrm{prog})$. \hfill $\square$

This provides a budget-aware justification for increasing $K$ under stagnation and decreasing $K$
under consistently successful progress. In the actual algorithm, $K$ is selected from the finite set
$\mathcal{K}=\{1,3,5,7\}$, so the continuous argument should be read as a design intuition: the
discrete scheduler approximates this regime-dependent ordering using recent archive-update feedback.
This argument does not establish the finite-set controller's optimality.

\paragraph{Why adaptive $K$ can outperform fixed $K$.}
Suppose the search spends a fraction $\rho$ of time in the progress regime and $1-\rho$ in the
stagnation regime. For any fixed candidate budget $K$, the average utility is
\begin{equation}
U_{\mathrm{fixed}}(K)
=
\rho \bigl[r_{\mathrm{prog}}(K)-\lambda c(K)\bigr]
+
(1-\rho)\bigl[r_{\mathrm{stag}}(K)-\lambda c(K)\bigr].
\label{eq:ufixed}
\end{equation}
Eq.~\eqref{eq:ufixed} shows that any fixed $K$ strategy is a \emph{static compromise} across
heterogeneous search regimes: if $K$ is too small, it under-explores during stagnation; if $K$ is too
large, it wastes budget during progress.

By contrast, a two-regime adaptive policy that uses $K_-$ during progress and $K_+$ during
stagnation has average utility
\begin{equation}
U_{\mathrm{adapt}}(K_-,K_+)
=
\rho \bigl[r_{\mathrm{prog}}(K_-)-\lambda c(K_-)\bigr]
+
(1-\rho)\bigl[r_{\mathrm{stag}}(K_+)-\lambda c(K_+)\bigr],
\qquad
K_- \le K_+.
\label{eq:uadapt}
\end{equation}
In the idealized case where the adaptive choices match the regime-specific optima,
$K_-=K^\star(\mathrm{prog})$ and $K_+=K^\star(\mathrm{stag})$, we have
\begin{equation}
U_{\mathrm{adapt}}\!\left(K^\star(\mathrm{prog}),K^\star(\mathrm{stag})\right)
\ge
U_{\mathrm{fixed}}(K),
\qquad \text{for any fixed admissible } K,
\label{eq:dominance}
\end{equation}
because each regime-specific term is optimized separately. Thus, fixed $K$ must use a single search
width for different search states, whereas adaptive $K$ can better match regime-specific budget needs
over time. Our island-local controller is a lightweight approximation to this principle: it increases
$K$ when recent proposal events fail to update the archive, decreases $K$ when every recent event
succeeds, and otherwise keeps $K$ unchanged to avoid oscillation.

\section{Prompt Comparison (Unified Diff) for Verbalized Sampling}
\label{appendix:prompt_diff_circle_packing}

We show a unified diff of the \emph{user prompt} change for \taskname{CirclePack\_26} (n=26).
The system prompt remains identical; only the requested output format and the number of rewrites differ.

\Needspace{12\baselineskip}
\begin{PromptDiffBox}{Unified diff of user prompt for \taskname{CirclePack\_26}}
--- Original user prompt (single rewrite)
+++ Our user prompt (Verbalized Sampling; K candidates with probabilities)
@@
 # Task
 Rewrite the program to improve its FITNESS SCORE.
 The system maintains diversity across these dimensions: {feature_dimensions}
 Different solutions with similar fitness but different features are valuable.
-Provide the complete new program code.
+Generate {k_candidates} candidate rewritten programs.

 IMPORTANT: Make sure your rewritten program maintains the same inputs and outputs
 as the original program, but with improved internal implementation.

-```{language}
-# Your rewritten program here
-```
+Return the responses in JSON format with the key: "responses" (list of dicts). Each dictionary must include:
+- "code": the complete program source code string only (no explanation or extra text). Do NOT wrap with any code fence like ```python```.
+- "probability": how likely this program would be (from 0.0 to 1.0).
+
+The probabilities across the {k_candidates} responses must sum to 1.0 (+/- 1e-3).
+Randomly sample the responses from the full distribution.
+
+Return ONLY the JSON object, with no additional explanations or text.
\end{PromptDiffBox}

\section{Task Details}
\label{appendix:task}

We use five executable mathematical construction tasks:
\taskname{Autocorr\_First} (\texttt{first\_\allowbreak{}autocorr\_\allowbreak{}ineq}),
\taskname{Autocorr\_Second} (\texttt{second\_\allowbreak{}autocorr\_\allowbreak{}ineq}),
\taskname{Uncertainty} (\texttt{uncertainty\_\allowbreak{}ineq}),
\taskname{CirclePack\_26} (\texttt{circle\_\allowbreak{}packing}, $n=26$), and
\taskname{Erdos\_Overlap} (\texttt{erdos\_\allowbreak{}min\_\allowbreak{}overlap}).
The three KernelBench Level 2 GPU-kernel tasks are:
\taskname{Matmul\_SwishGN} (\texttt{37\_\allowbreak{}Matmul\_\allowbreak{}Swish\_\allowbreak{}Sum\_\allowbreak{}GroupNorm}),
\taskname{Conv3d\_Mish} (\texttt{47\_\allowbreak{}Conv3d\_\allowbreak{}Mish\_\allowbreak{}Tanh}), and
\taskname{Conv2d\_BN} (\texttt{73\_\allowbreak{}Conv2d\_\allowbreak{}BatchNorm\_\allowbreak{}Scaling}).
Parentheses give the benchmark identifiers corresponding to the task aliases.
Each task provides an automatic evaluator that (i) verifies feasibility and hard constraints and
(ii) returns a scalar score. For inequality-style tasks, candidate constructions typically certify
either an upper bound or a lower bound on a target constant; accordingly, some tasks are minimized
(tighten an upper bound, $\downarrow$) while others are maximized (raise a lower bound or improve an objective, $\uparrow$).
The KernelBench L2 tasks are GPU operator-sequence optimization tasks, where Level 2 consists of
multi-operator neural-network workloads that expose opportunities for kernel fusion; in our setting,
their scalar score is the speedup over the PyTorch reference runtime, and is therefore maximized.

\paragraph{\taskname{Autocorr\_First} ($\downarrow$).}
For a function $f:\mathbb{R}\to\mathbb{R}$, define its autoconvolution
$(f*f)(t)=\int_{\mathbb{R}} f(t-x)f(x)\,dx$.
The task concerns an inequality of the form
\[
\max_{-1/2\le t\le 1/2} (f*f)(t)\ \ge\ C_1\left(\int_{-1/4}^{1/4} f(x)\,dx\right)^2
\]
required to hold for all non-negative $f$.
A feasible construction (implemented as code, typically specifying a structured/discretized non-negative function)
induces a computable ratio that certifies an \emph{upper bound} on the optimal constant $C_1$.
The evaluator returns this certified bound, so the objective is to minimize it.

\paragraph{\taskname{Autocorr\_Second} ($\uparrow$).}
This task studies an autocorrelation inequality involving norms of $f*f$, e.g.,
\[
\|f*f\|_2^2 \ \le\ C_2\,\|f*f\|_1\,\|f*f\|_\infty
\]
for all non-negative $f$.
Any feasible $f$ yields a normalized quantity
$\|f*f\|_2^2 /(\|f*f\|_1\|f*f\|_\infty)$ that certifies a \emph{lower bound} on $C_2$.
Candidate programs specify a non-negative function family; the evaluator computes the norms of $f*f$
and returns the certified lower bound, which we aim to maximize.

\paragraph{\taskname{Uncertainty} ($\downarrow$).}
Let $\hat f(\xi)=\int_{\mathbb{R}} f(x)e^{-2\pi i x\xi}\,dx$ be the Fourier transform, and define
$A(f)=\inf\{r>0:\ f(x)\ge 0\ \text{for all } |x|\ge r\}$.
The target inequality is of the form
\[
A(f)\,A(\hat f)\ \ge\ C_4,
\]
under feasibility conditions on $f$ (e.g., evenness and sign constraints at the origin).
A feasible test function $f$ produces a numeric value $A(f)A(\hat f)$ that certifies an \emph{upper bound}
on the best possible constant $C_4$.
Thus, the evaluator returns this certificate and the optimization goal is to minimize it.
In practice, candidate programs parameterize structured even functions and the evaluator checks feasibility
and estimates $A(\cdot)$ for both $f$ and $\hat f$.

\paragraph{\taskname{CirclePack\_26} (n=26) ($\uparrow$).}
The goal is to place $n=26$ pairwise disjoint circles inside the unit square to maximize the sum of radii.
A candidate program outputs a packing configuration (e.g., circle centers and radii, or an equivalent parametrization).
The evaluator checks containment within the square and non-overlap constraints, then returns the objective value
(sum of radii) to be maximized.

\paragraph{\taskname{Erdos\_Overlap} ($\downarrow$).}
This task targets the minimum overlap problem via an inequality-based certificate.
Candidate programs construct a normalized function (commonly represented as a bounded step function),
and the evaluator computes an overlap-related objective that involves maximizing a shifted interaction term
over a range of shifts.
A concrete construction yields a numeric certificate that upper-bounds the target constant;
therefore, the optimization direction is to minimize the evaluator's returned value.

\paragraph{\taskname{Matmul\_SwishGN} ($\uparrow$).}
This task consists of a dense linear projection followed by Swish activation,
bias addition, and group normalization:
\[
x \mapsto \mathrm{GroupNorm}\bigl(\mathrm{Swish}(xW^\top+b_{\mathrm{lin}})+b_{\mathrm{add}}\bigr),
\quad
\mathrm{Swish}(z)=z\sigma(z).
\]
The reference workload uses a batch size of $32768$, input dimension $1024$, output dimension $4096$,
and $64$ groups for GroupNorm.
The task combines a large matrix multiplication with elementwise activation, vector addition,
and normalization, making it representative of fused post-processing around a compute-intensive
GEMM-like operator.
The score is the achieved speedup over the PyTorch reference implementation, so the objective is to
maximize it.

\paragraph{\taskname{Conv3d\_Mish} ($\uparrow$).}
A three-dimensional convolution is followed by two nonlinear activations:
\[
x \mapsto \tanh(\mathrm{Mish}(\mathrm{Conv3d}(x))).
\]
The input tensor has batch size $16$, $32$ input channels, and spatial size
$(D,H,W)=(32,64,64)$; the convolution produces $64$ output channels with kernel size $3$.
The workload represents a convolution-dominated operator sequence with pointwise nonlinear
post-processing, where optimization opportunities arise from reducing the overhead and memory traffic
of separately materialized activation stages.
The evaluator returns the speedup relative to the PyTorch reference runtime, which we aim to maximize.

\paragraph{\taskname{Conv2d\_BN} ($\uparrow$).}
The task combines 2D convolution, batch normalization, and constant scaling:
\[
x \mapsto \alpha\,\mathrm{BatchNorm2d}(\mathrm{Conv2d}(x)),
\qquad \alpha=2.0.
\]
The reference workload uses batch size $128$, $8$ input channels, image size $(128,128)$,
$64$ output channels, and convolution kernel size $3$.
It captures a common convolutional-network pattern in which normalization and affine rescaling follow
a spatial convolution, exposing opportunities to fuse or streamline the post-convolution computation.
The scalar objective is speedup over the PyTorch reference implementation, which we maximize.

\section{Protocol Details}
\label{appendix:budget_protocol}

\paragraph{Comparison scope.}
For each main Gemini method--task--backend configuration, we run five repeated runs under the
protocol described in Section~\ref{subsec:protocol}. Unless otherwise stated, warm-start initialization is disabled in
the main controlled comparisons and evaluated separately in the initialization study. All controlled
reruns within a task--backend block share the same executable evaluator, LLM backend,
cost-accounting rule, and stopping protocol. \sysname{AlphaEvolve} is included only as a reported
reference and is excluded from all within-block rankings because no public implementation is
available for matched-protocol reruns.

\paragraph{Cost accounting.}
For each run, we compute API cost from logged input and output tokens using the corresponding
provider pricing for the selected backend, including provider-side prompt/prefix caching.
The same accounting rule is applied to all controlled
methods within each task--backend block. API cost refers to model-call cost only; local evaluator
compute is not included in the API-cost budget. All model calls are counted, including retry or
re-query calls caused by malformed outputs, insufficient returned candidates, parsing failures, or
other format-related issues. Their input/output tokens are included in the accumulated API cost, and
the calls are included when reporting LLM invocation counts.

\paragraph{Matched API-cost budgets.}
For each task and backend, we first run \sysname{TurboEvolve} to a predefined evaluated-program
horizon and use its average API cost over five runs as the matched cost budget for controlled
baselines. An evaluated program is a candidate submitted to the executable evaluator; under VS, one
LLM invocation may produce multiple evaluated candidates. The reference horizon is 800 evaluated
programs for the mathematical/algorithmic tasks and 100 evaluated programs for the KernelBench
tasks. Each controlled baseline is stopped once its accumulated API cost reaches or exceeds the
corresponding task-specific cost budget. Since API cost is known only after a model call is completed,
the final call of a run may cause the accumulated cost to slightly exceed the target budget; after this
point, no further LLM calls are made. The reported score is the best score observed within the
resulting run budget.

\paragraph{Equal evaluated-program horizon diagnostic.}
In addition to matched API-cost comparisons, we run the \sysname{OpenEvolve}-backbone diagnostic
to the same evaluated-program horizon as \sysname{TurboEvolve}. This diagnostic does not stop
\sysname{OpenEvolve} at \sysname{TurboEvolve}'s API-cost budget; instead, it measures how many LLM
calls and how much API cost are required by the single-descendant loop to evaluate the same number
of programs. We compare endpoint scores, LLM calls, and API costs at this matched program horizon.

\paragraph{Ranking and reporting.}
Best and second-best annotations in the main result table are computed only among controlled reruns
within the same backend block. Table~\ref{tab:main_results} reports best-over-runs results: for each controlled method
and task, we report the strongest score found across five runs under the matched API-cost budget.
Robustness figures provide the complementary run-level view, reporting the mean and standard
deviation of final per-run best scores over five runs. Trajectory and mechanism-diagnostic figures
use run-level aggregation and summarize variability by standard deviation or IQR as specified in
their captions.

\paragraph{Failed evaluations.}
Compilation errors, runtime exceptions, timeouts, infeasible outputs, and hard-constraint violations
are counted once a candidate enters the evaluation pipeline. Such candidates do not update the
archive, but they are counted as evaluated programs for the corresponding evaluated-program horizon.
If a model call fails before producing any candidate that can enter the evaluator, the corresponding
API cost and LLM invocation are still counted, but no evaluated-program count is added for that call.

\paragraph{Computing environment.}
The main Gemini experiments used a server equipped with two AMD EPYC 7Y43 48-Core processors,
providing 96 physical CPU cores and 192 threads, with approximately 1.0 TiB of system memory.
The server contains 8 NVIDIA GeForce RTX 4090 GPUs (48 GB variants), each with 49,140 MiB of reported memory. The
NVIDIA driver version was 570.211.01, with CUDA 12.8 reported by \texttt{nvidia-smi}. The
operating system was Ubuntu 22.04.5 LTS with Linux kernel 6.8.0-106-generic.
The additional local-inference experiments used four NVIDIA H200 GPUs
(Appendix~\ref{appendix:local_inference}).

\newpage
\section{Additional Robustness Results}
\label{appendix:robustness_g3}

\begin{figure}[H]
    \centering
    \includegraphics[width=\textwidth]{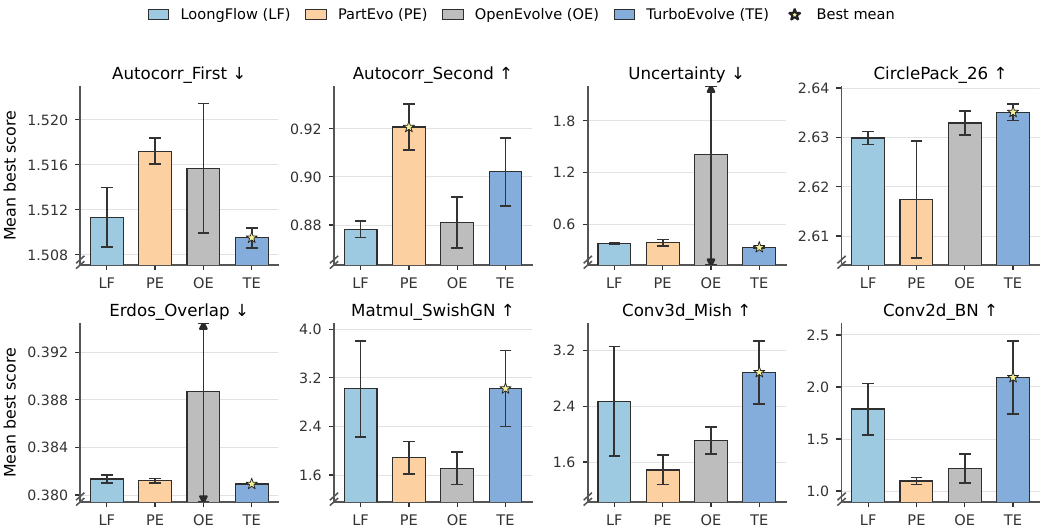}
    \caption{Run-to-run consistency under Gemini-3-Flash. Bars show the mean of final per-run best scores over five runs, and error bars show standard deviation.}
    \label{fig:robustness_g3}
\end{figure}

\section{Additional Backbone-Paired Efficiency Trajectories}
\label{appendix:efficiency_trajectories}

Figure~\ref{fig:backbone_efficiency}(A) in the main text reports the full endpoint scores and
resource measurements at equal evaluated-program horizons. Averaging the percentage reductions
equally over the eight tasks within each backend gives 73.1\% fewer calls and 63.5\% lower API cost
for Gemini-2.5-Flash, and 73.5\% fewer calls and 79.1\% lower API cost for Gemini-3-Flash.
\sysname{TurboEvolve} has the better endpoint score on seven of eight and six of eight tasks, respectively.

\begin{figure}[p]
\centering
\includegraphics[width=0.47\textwidth]{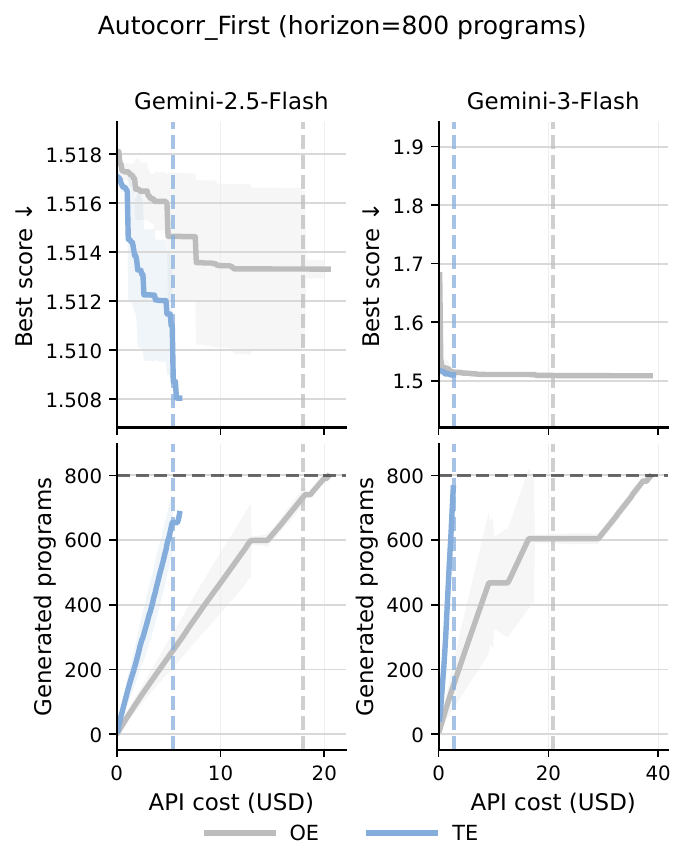}\hfill
\includegraphics[width=0.47\textwidth]{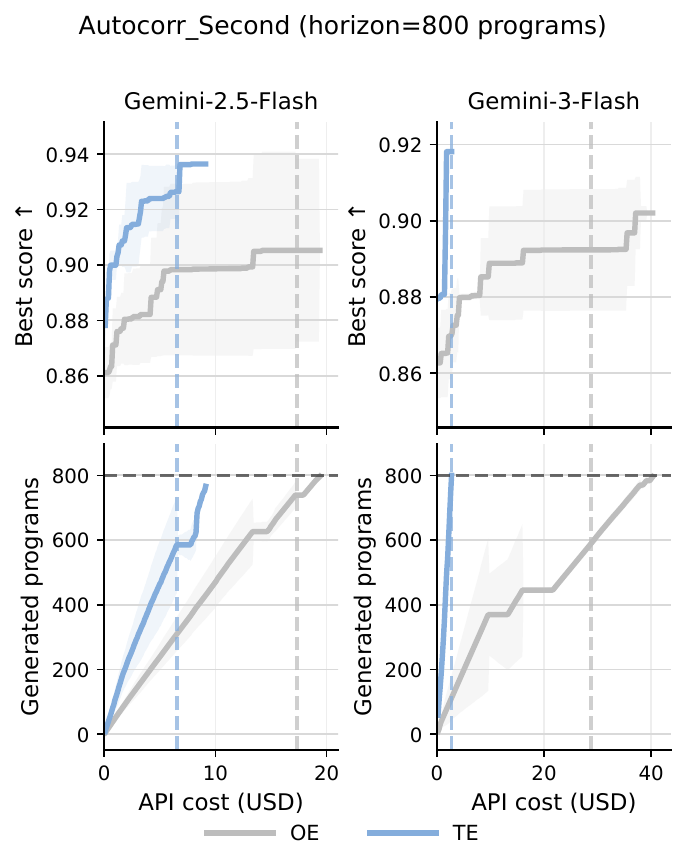}\\[6pt]
\includegraphics[width=0.47\textwidth]{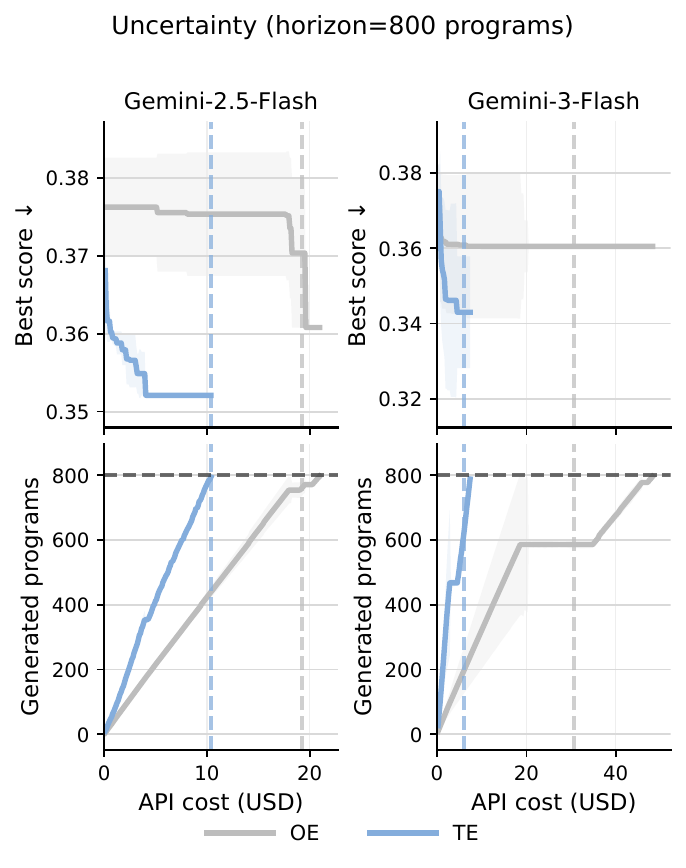}\hfill
\includegraphics[width=0.47\textwidth]{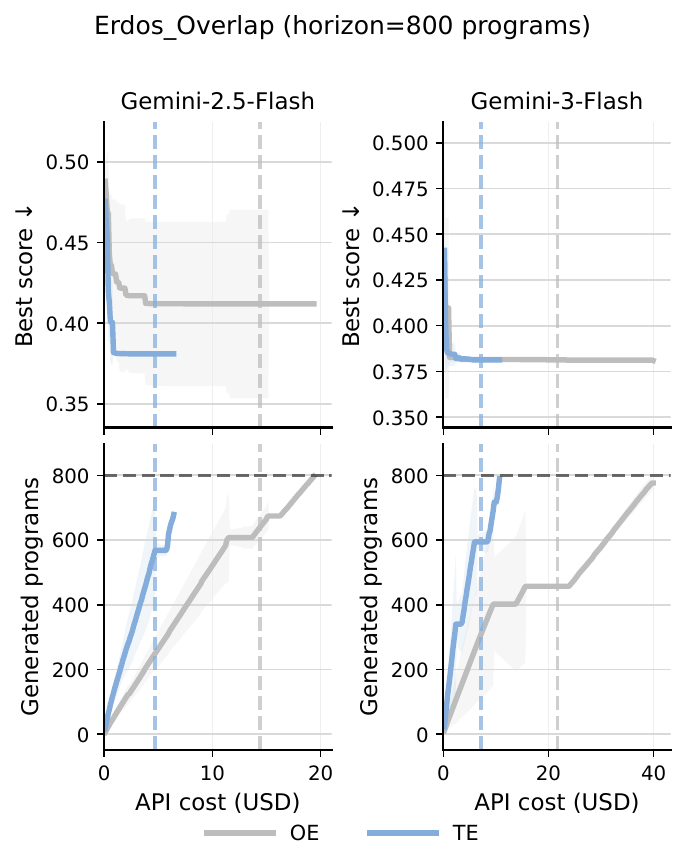}
\caption{Additional backbone-paired efficiency trajectories: Mathematical tasks. Each task panel shows best-so-far score and generated programs versus API cost for \sysname{OpenEvolve} and \sysname{TurboEvolve}. Columns within each panel correspond to Gemini-2.5-Flash and Gemini-3-Flash.}
\label{fig:efficiency_trajectory_gallery_b}
\end{figure}

\begin{figure}[p]
\centering
\includegraphics[width=0.47\textwidth]{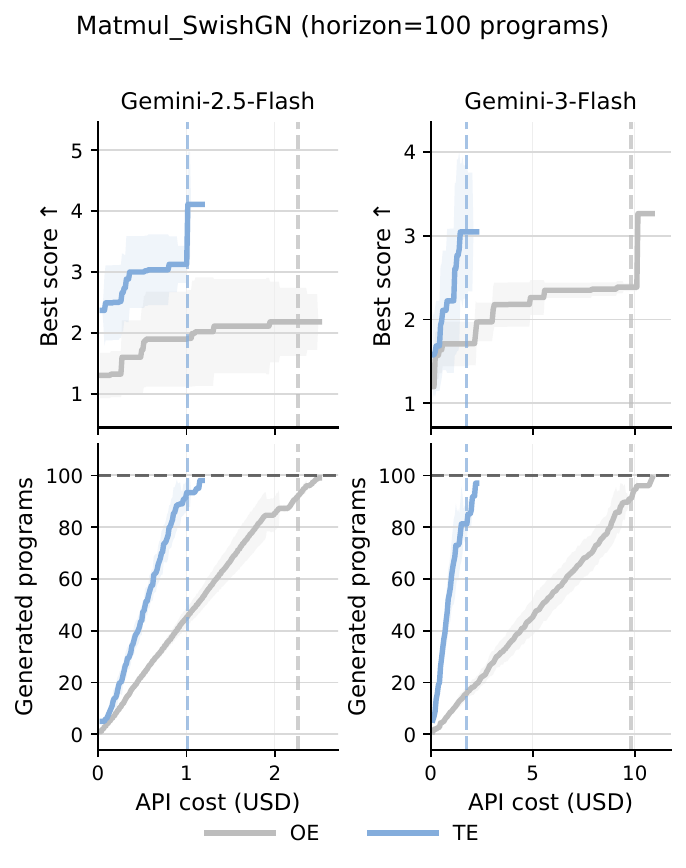}\hfill
\includegraphics[width=0.47\textwidth]{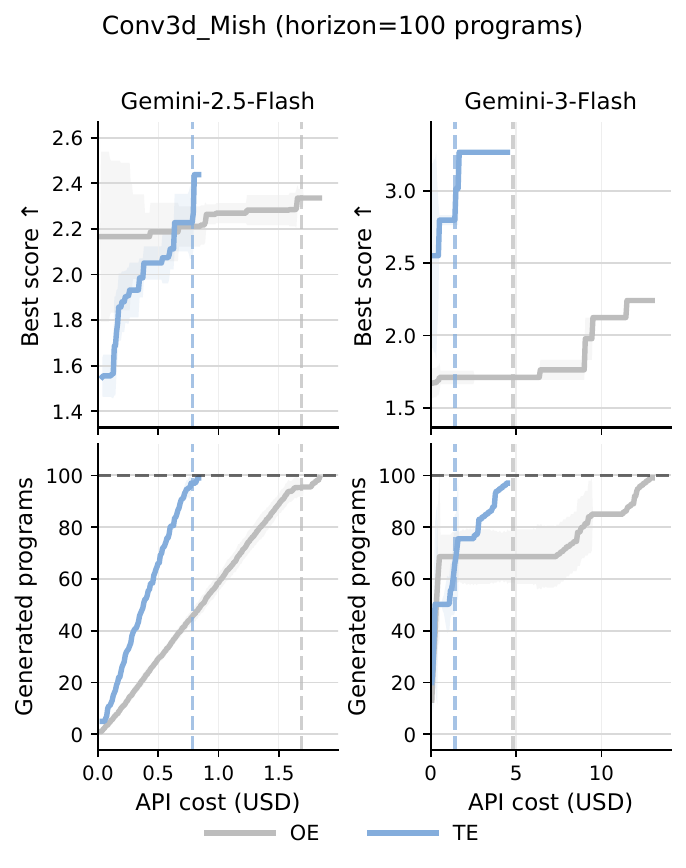}\\[6pt]
\includegraphics[width=0.47\textwidth]{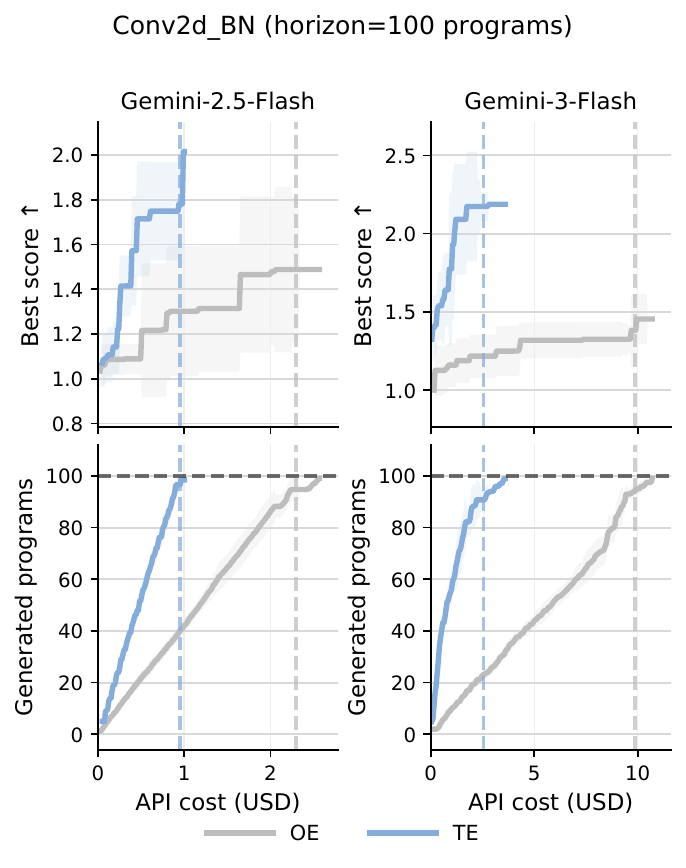}
\caption{Additional backbone-paired efficiency trajectories: KernelBench tasks. Each task panel shows best-so-far score and generated programs versus API cost for \sysname{OpenEvolve} and \sysname{TurboEvolve}. Columns within each panel correspond to Gemini-2.5-Flash and Gemini-3-Flash.}
\label{fig:efficiency_trajectory_gallery_d}
\end{figure}

Figures~\ref{fig:efficiency_trajectory_gallery_b}--\ref{fig:efficiency_trajectory_gallery_d}
provide the remaining backbone-paired efficiency trajectories omitted from the main text.
All panels follow the same format as Fig.~\ref{fig:backbone_efficiency}(B): columns correspond to
Gemini-2.5-Flash and Gemini-3-Flash, while rows show best-so-far score and generated programs
versus API cost. The \taskname{CirclePack\_26} trajectory is shown in Fig.~\ref{fig:backbone_efficiency}(B).


\FloatBarrier
\section{Responsible Use of Seed-Pool Initialization}
\label{appendix:seed_pool_responsible_use}

Seed-pool initialization is intended to make LLM-driven program evolution more efficient and reproducible when prior runs, external baselines, or reusable program libraries are available. 
Instead of restarting every island from the same seed or an unstructured pool, \sysname{TurboEvolve} allocates seed programs across islands through representation-based clustering, controlled cross-cluster mixing, and elitist preservation. 
This design helps preserve distinct solution modes, share useful motifs across islands, and reduce wasted search budget during the warm-start phase.

At the same time, executable seed pools should be treated carefully when they come from external or community-shared sources. 
Untrusted seed programs may contain backdoors, poisoned examples, license conflicts, or other supply-chain risks that could be propagated or refined by the evolutionary loop. 
For future systems that rely on external seed libraries, we recommend complementary safeguards, including provenance auditing, license verification, static and dynamic security scanning, sandboxed execution, and human review in safety-critical settings. 
These measures help retain the efficiency and diversity benefits of warm-start initialization while reducing risks associated with untrusted executable seeds.

\section{Additional Mechanism Diagnostics}
\label{appendix:more_analysis}

The main experiments report task-level performance and efficiency; here we provide additional
mechanism-oriented diagnostics to contextualize how \sysname{TurboEvolve} behaves during search.
We focus on two core ingredients: (i) multi-candidate generation via Verbalized Sampling (VS) and
(ii) online budget allocation through adaptive candidate count \(K\). We use
\taskname{CirclePack\_26} (\(n{=}26\)) as a representative case study. 
Fig.~\ref{fig:analysis_1} studies the marginal utility of retaining longer returned prefixes within
large-\(K\) events via within-event prefix-\(m\) replay, avoiding direct comparisons between
independently scheduled \(K\) values. Fig.~\ref{fig:analysis_2} profiles large-\(K\) calls by the
returned VS rank to examine quality--diversity roles within one invocation. 
Fig.~\ref{fig:analysis_3} examines how event outcomes vary with both \(K\) and a pre-generation
context signal, namely the parent--inspiration distance. Finally, Fig.~\ref{fig:analysis_4} reports a
quality--diversity (QD) archive footprint to complement global-best learning curves with coverage and per-cell
quality diagnostics. These analyses are descriptive rather than causal; unless otherwise stated,
statistics are aggregated at the run level and summarized across runs by the median with an IQR band.

\paragraph{Logging and definitions.}
In this diagnostic appendix, an \emph{event} refers to one LLM invocation, distinct from a
scheduler proposal event, which groups retry attempts into one window observation
(Section~\ref{subsec:adaptive_k}). For every diagnostic event, we log the island id and iteration
index, the programs used in the prompt (the primary parent and inspirations), the chosen candidate
count \(K\in\{1,3,5,7\}\), and per-candidate outcomes for the \(K\) generated programs: returned VS
rank, executability/validity, objective score, and an improvement signal \(\Delta\), defined relative to
the incumbent used for selection in that event. We also record the archive cell of each accepted
candidate, which enables diversity proxies such as archive-cell distance to the primary parent. In
addition, we snapshot the archive at predefined evaluated-program budgets to compute
\emph{coverage} (the number of occupied cells) and \emph{cell quality} (the mean best score per
occupied cell).

To summarize prompt ``signal strength'', we group events by \emph{inspiration tier}, defined by the
maximum inspiration score in the prompt and discretized into five within-run quantile bins
(Tier~1 lowest, Tier~5 highest). This stratification is used only for analysis and does not affect the
algorithm.

\subsection{Within-event prefix-\texorpdfstring{$m$}{m} utility under adaptive \texorpdfstring{$K$}{K}}
\label{sec:more_counterfactual}

Figure~\ref{fig:analysis_1} in the main text presents this replay. We detail its construction and interpretation below.

Because \(K\) is chosen online, events with different \(K\) values are not directly comparable:
larger \(K\) is more likely to occur in phases with weaker recent archive-update feedback.
The fixed \(K\) ablation in Section~\ref{sec:adaptive_k_ablation} evaluates the full-run effect of adaptive scheduling against
static candidate-count choices. Here, we ask a complementary within-call question: when the scheduler
has already selected a large \(K\), do the additional returned candidates provide useful marginal signal
within the same event?

To answer this, we perform a \emph{within-event replay}. For each \(K{=}7\) event, we hold the
generated candidate set fixed and recompute two event-local metrics as a function of \(m\le 7\):
(i) \emph{improvement coverage}, whether any of the first \(m\) candidates in the returned VS order
yields a valid improvement, and (ii) \emph{best score change}, \(\max_{j\le m}\Delta_j\).
This replay avoids comparing independently scheduled \(K\) values and instead isolates the marginal
utility of retaining longer prefixes within the same returned candidate set.

The gain is more pronounced for higher inspiration tiers, suggesting that additional VS
candidates provide larger marginal benefit when the prompt context contains stronger improvement
signals.

\subsection{Large-\texorpdfstring{$K$}{K} rank profile within a single call}
\label{sec:more_rank_profile}

\begin{figure}[ht]
    \centering
    \includegraphics[width=\linewidth]{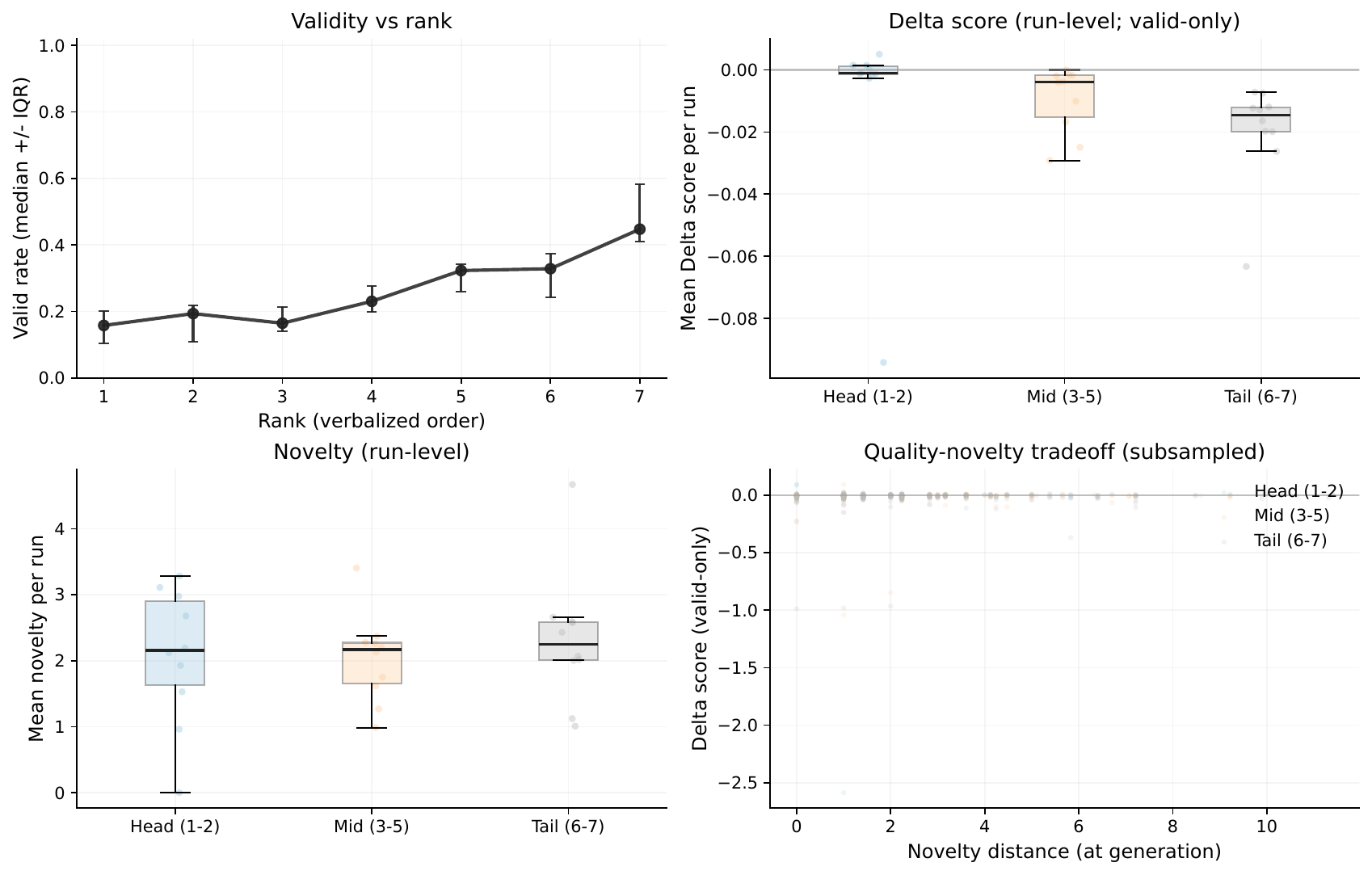}
    \caption{\textbf{Rank-wise quality and diversity under large \(K\).}
    We profile candidates \emph{within the same LLM call} by their rank in the returned VS order.
    We report validity rate, valid-only score improvement \(\Delta\), and a diversity proxy
    (archive cell-distance to the primary parent), aggregated at the run level with median and IQR.
    For panels that compare rank groups, we summarize ranks into head (1--2), mid (3--5), and
    tail (6--7) using the same \(K{=}7\) events.}
    \label{fig:analysis_2}
\end{figure}

The within-event replay in \S\ref{sec:more_counterfactual} quantifies the marginal value of retaining
longer prefixes from a large-\(K\) candidate set, but it does not reveal how different returned ranks
contribute within a single call. We therefore profile \(K{=}7\) events by the rank of each candidate in
the returned VS order. For each rank or rank group, we summarize three run-level quantities:
validity rate, valid-only score improvement \(\Delta\), and a diversity proxy given by archive
cell-distance to the primary parent.

Figure~\ref{fig:analysis_2} shows that the returned ranks do not form a simple monotonic quality
ordering. Head candidates have the strongest valid-only immediate improvements, but are not
necessarily the most valid; tail candidates are more often valid in our runs, but their valid-only
improvements are weaker on average. In parallel, the diversity proxy shows that later candidates,
especially the tail group, provide alternatives that are farther from the primary parent in archive space.
The candidate-level quality--novelty scatter further suggests that novelty alone is not a reliable
indicator of immediate score gain: high-novelty candidates do not consistently yield large positive
\(\Delta\) in the same iteration. Taken together, these patterns support a complementary-role view of
large-\(K\) VS generation: early candidates provide stronger direct-improvement signals when valid,
whereas later candidates broaden the valid and diverse alternatives available for archive selection.

\subsection{Distance--\texorpdfstring{$K$}{K} interaction under adaptive scheduling}
\label{sec:more_distance_k}

\begin{figure}[htbp]
    \centering
    \includegraphics[width=\linewidth]{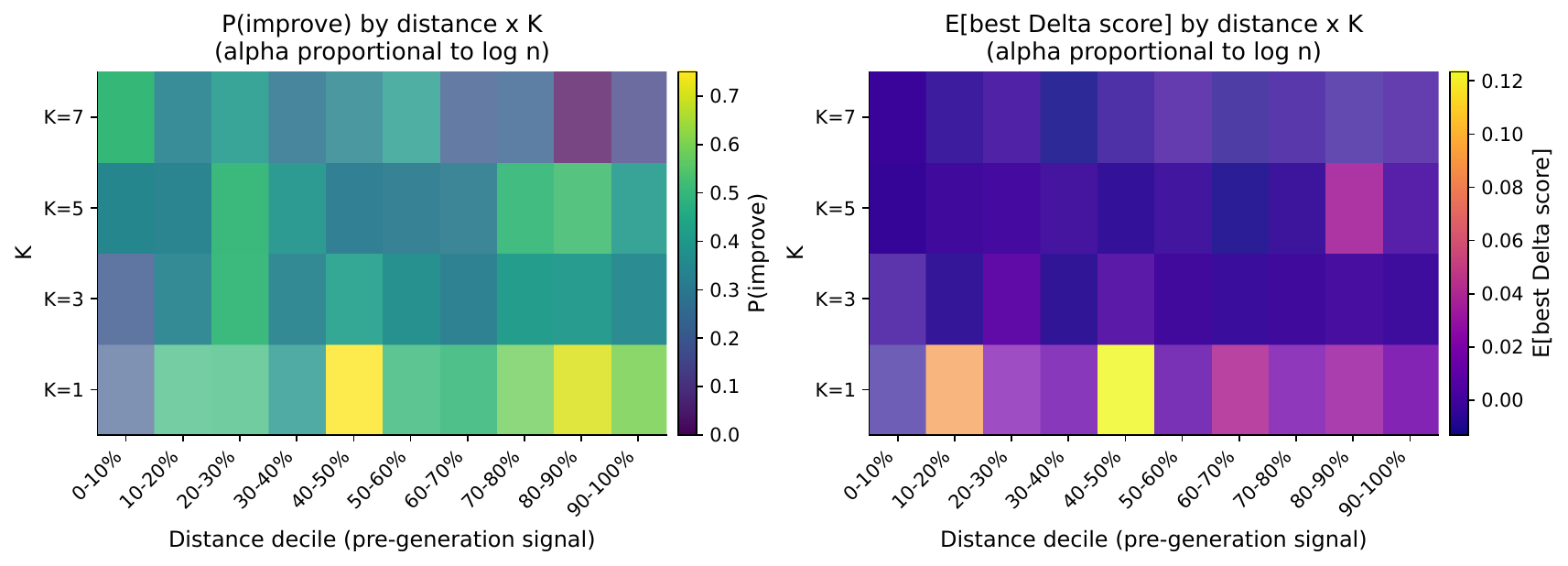}
    \caption{\textbf{Distance--\(K\) interaction under adaptive scheduling.}
    Heatmaps report (A) \(P(\mathrm{improve})\) and (B) \(\mathbb{E}[\Delta_{\mathrm{best}}]\) over
    pre-generation distance deciles and \(K\in\{1,3,5,7\}\).
    Events are first aggregated within each run on the \((\text{distance},K)\) grid and then summarized
    across runs; cell opacity reflects the number of events in each bin.}
    \label{fig:analysis_3}
\end{figure}

Because \(K\) is adjusted online, events with different \(K\) values are not exchangeable:
larger \(K\) is more likely to be selected in phases with weaker recent archive-update feedback.
A naive across-\(K\) comparison would therefore conflate \emph{candidate count} with the difficulty
of the search phase.
We use Fig.~\ref{fig:analysis_3} as a contextual diagnostic rather than a causal estimate:
instead of asking whether one \(K\) is globally better, we ask whether the payoff of a given \(K\)
varies with a pre-generation context signal.

For each event, we compute a \emph{pre-generation} distance between the primary parent and the
inspiration source in a fixed representation space, and discretize this distance into 10 within-run
deciles.
We then evaluate two event-level outcomes on the \((\text{distance},K)\) grid:
(i) \(P(\mathrm{improve})\), whether at least one generated candidate achieves \(\Delta>0\); and
(ii) \(\mathbb{E}[\Delta_{\mathrm{best}}]\), the best improvement within the event relative to the
incumbent used for selection.
To avoid treating multiple candidates from the same event as independent samples, we aggregate
events within each \((\text{distance},K)\) cell at the run level before summarizing across runs.

Figure~\ref{fig:analysis_3} shows that event outcomes vary substantially with the pre-generation
distance, and that this dependence is not captured by \(K\) alone. Very near inspirations often provide
limited room for improvement, whereas very distant inspirations can be harder to translate into
feasible, constraint-satisfying gains. Intermediate distances more often balance novelty and feasibility,
although the pattern is not strictly monotonic and differs across \(K\) values. This supports using
distance as a lightweight context indicator for future scheduling and inspiration-selection policies.
Importantly, the analysis remains descriptive: it suggests that the utility of expanding \(K\) is
context-dependent, but does not claim that the current scheduler is optimal or that distance alone is
sufficient for deciding \(K\).

\subsection{QD footprint: coverage and per-cell quality}
\label{sec:more_qd}

\begin{figure}[htbp]
    \centering
    \includegraphics[width=\linewidth]{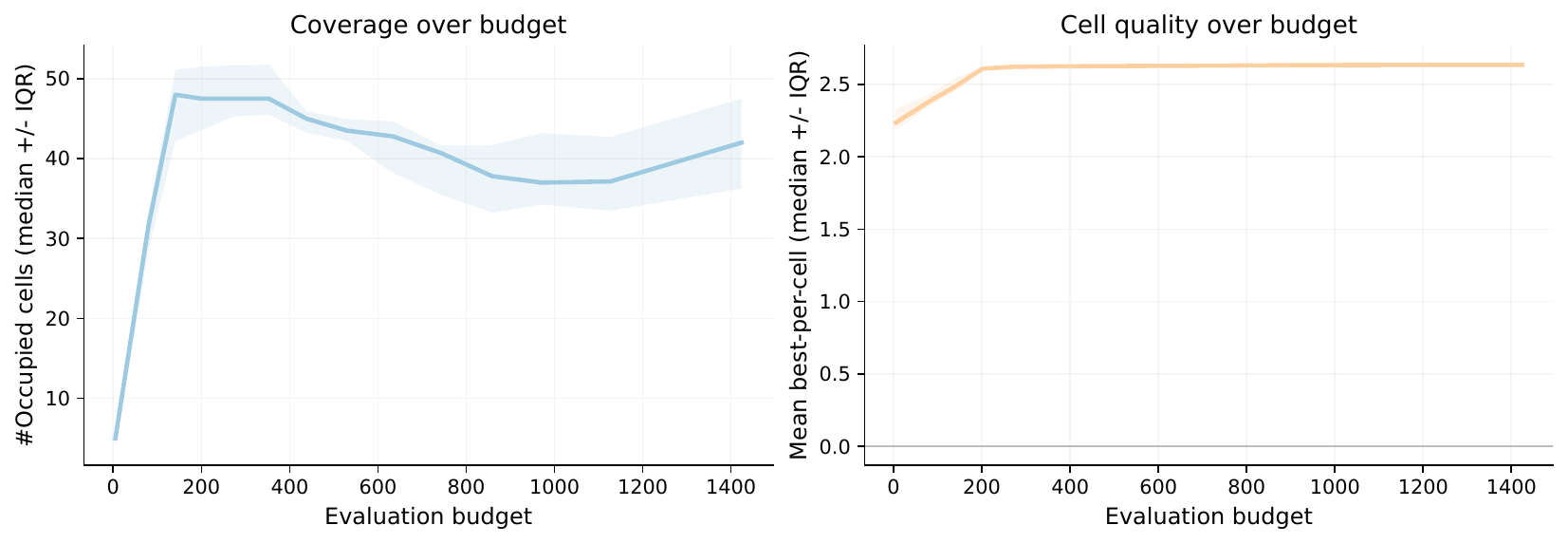}
    \caption{\textbf{QD footprint over evaluation budget.}
    We report archive \emph{coverage} (number of occupied cells) and \emph{cell quality}
    (mean best score per occupied cell) as functions of evaluation budget for \sysname{TurboEvolve}.
    Each run is interpolated onto a shared budget grid; lines show the median across runs and
    bands indicate IQR.}
    \label{fig:analysis_4}
\end{figure}

Main learning curves track the global best score versus evaluation budget, but archive-based methods
also maintain a population of solutions across map cells. We therefore include a QD-style footprint as
an archive-level diagnostic, summarizing both \emph{coverage} and \emph{per-cell quality} over the
evaluation budget.

Figure~\ref{fig:analysis_4} shows a two-stage archive dynamic. In the early phase,
\sysname{TurboEvolve} rapidly expands coverage, indicating that multi-candidate proposal can populate
many archive cells within a small evaluation budget. After this initial expansion, coverage fluctuates
rather than increasing monotonically, while mean best-per-cell quality remains stable. This suggests
that subsequent search is not merely adding more occupied cells, but also refining and maintaining
quality within the existing archive footprint. Thus, the gains of \sysname{TurboEvolve} are not only
reflected in the single best trajectory; they are also consistent with a search process that preserves a
broad archive while sustaining cell-level quality. This analysis is descriptive and complements the
best-score and efficiency results in the main text.

\input{sections/additional_experiments}

\end{document}

%% file: tabs/main_result.tex
\begin{table}[t]
\centering
\caption{Best-over-runs scores under matched API-cost budgets. Each controlled configuration is
run five times. Arrows indicate optimization direction; bold and underline mark the best and
second-best controlled results within each backend block. \sysname{AlphaEvolve} values are reported references
with a different backend and are excluded from rankings.}
\label{tab:main_results}
\par\vspace{\baselineskip}
\begin{tabularx}{\textwidth}{@{}l@{\hspace{0.5em}}|*{8}{>{\centering\arraybackslash}X@{}}}
\hline
\textbf{Method} &
\makecell[t]{\taskname{Autocorr\_}\\\taskname{First}$\downarrow$} &
\makecell[t]{\taskname{Autocorr\_}\\\taskname{Second}$\uparrow$} &
\makecell[t]{\taskname{Uncer-}\\\taskname{tainty}$\downarrow$} &
\makecell[t]{\taskname{Circle}\\\taskname{Pack\_26}$\uparrow$} &
\makecell[t]{\taskname{Erdos\_}\\\taskname{Overlap}$\downarrow$} &
\makecell[t]{\taskname{Matmul\_}\\\taskname{SwishGN}$\uparrow$} &
\makecell[t]{\taskname{Conv3d\_}\\\taskname{Mish}$\uparrow$} &
\makecell[t]{\taskname{Conv2d\_}\\\taskname{BN}$\uparrow$} \\
\hline
& \multicolumn{8}{c@{}}{Gemini-2.0-Flash/Pro}\\
\hline
\sysname{AlphaEvolve} & 1.505294 & 0.896280 & 0.321587 & 2.635863 & 0.380923 & -- & -- & -- \\
\hline
& \multicolumn{8}{c@{}}{Gemini-2.5-Flash}\\
\hline
\sysname{LoongFlow} & 1.513806 & 0.910976 & 0.358684 & 2.570675 & 0.381078 & 1.852214 & 2.207348 & 1.198789 \\
\sysname{PartEvo} & 1.510948 & 0.925310 & \second{0.352101} & 2.377295 & \second{0.381039} & 1.356195 & 1.432049 & 1.031080 \\
\sysname{OpenEvolve} & \second{1.509724} & \second{0.934569} & 0.358149 & \second{2.630972} & 0.381420 & \second{1.894889} & \second{2.210669} & \second{1.301967} \\
\sysname{TurboEvolve} & \best{1.507259} & \best{0.936662} & \best{0.352099} & \best{2.635983} & \best{0.380937} & \best{3.470687} & \best{2.255819} & \best{1.748876} \\
\hline
& \multicolumn{8}{c@{}}{Gemini-3-Flash}\\
\hline
\sysname{LoongFlow} & 1.508859 & 0.881876 & 0.365261 & 2.631350 & 0.381000 & \second{3.018998} & \second{2.470487} & \second{1.787289} \\
\sysname{PartEvo} & 1.515044 & \best{0.931462} & \second{0.332813} & 2.635328 & 0.380985 & 1.883057 & 1.485721 & 1.095771 \\
\sysname{OpenEvolve} & \second{1.508785} & 0.899975 & 0.334376 & \second{2.635896} & \second{0.380910} & 2.113726 & 1.910829 & 1.619962 \\
\sysname{TurboEvolve} & \best{1.508157} & \second{0.930484} & \best{0.328799} & \best{2.635984} & \best{0.380893} & \best{3.022276} & \best{2.885762} & \best{2.091736} \\
\hline
\end{tabularx}
\par\vspace{\baselineskip}
\end{table}

%% file: tabs/efficiency_full.tex
\begin{tabular*}{\linewidth}{@{}l@{\extracolsep{\fill}}c@{}c@{}c@{}}
\hline
\textbf{Task} & \textbf{Best score} & \textbf{LLM calls} & \textbf{API cost (\$)} \\
 & \textbf{OE\,/\,TE} & \textbf{OE\,/\,TE} & \textbf{OE\,/\,TE} \\
\hline
\multicolumn{4}{c}{Gemini-2.5-Flash}\\
\hline

\taskname{Autocorr\_First} $\downarrow$  
& 1.51162\,/\,\textbf{1.50997} 
& 800\,/\,\textbf{204} 
& 16.49\,/\,\textbf{5.39} \\
\taskname{Autocorr\_Second} $\uparrow$ 
& 0.90251\,/\,\textbf{0.91688} 
& 800\,/\,\textbf{153.2} 
& 16.42\,/\,\textbf{6.53} \\
\taskname{Uncertainty} $\downarrow$ 
& 0.36277\,/\,\textbf{0.35363} 
& 800\,/\,\textbf{193.8} 
& 19.64\,/\,\textbf{10.31} \\
\taskname{CirclePack\_26} $\uparrow$  
& 2.63129\,/\,\textbf{2.63585} 
& 800\,/\,\textbf{210.4} 
& 22.51\,/\,\textbf{1.12} \\
\taskname{Erdos\_Overlap} $\downarrow$   
& 0.40553\,/\,\textbf{0.38096} 
& 800\,/\,\textbf{242.8} 
& 15.50\,/\,\textbf{4.68} \\
\taskname{Matmul\_SwishGN} $\uparrow$     
& 2.28507\,/\,\textbf{3.47069} 
& 100\,/\,\textbf{31} 
& 2.26\,/\,\textbf{1.01} \\
\taskname{Conv3d\_Mish} $\uparrow$      
& \textbf{2.32806}\,/\,2.25582 
& 100\,/\,\textbf{30.4} 
& 1.69\,/\,\textbf{0.78} \\
\taskname{Conv2d\_BN} $\uparrow$        
& 1.47913\,/\,\textbf{1.74888} 
& 100\,/\,\textbf{28.2} 
& 2.30\,/\,\textbf{0.95} \\
\hline
\multicolumn{4}{c}{Gemini-3-Flash}\\
\hline
\taskname{Autocorr\_First} $\downarrow$  
& \textbf{1.50938}\,/\,1.50951 
& 800\,/\,\textbf{191.6} 
& 13.97\,/\,\textbf{2.00} \\
\taskname{Autocorr\_Second} $\uparrow$ 
& 0.88950\,/\,\textbf{0.89913} 
& 800\,/\,\textbf{215} 
& 12.86\,/\,\textbf{1.96} \\
\taskname{Uncertainty} $\downarrow$ 
& 0.37678\,/\,\textbf{0.33055} 
& 800\,/\,\textbf{151.8} 
& 19.56\,/\,\textbf{2.95} \\
\taskname{CirclePack\_26} $\uparrow$  
& \textbf{2.63521}\,/\,2.63478 
& 800\,/\,\textbf{127} 
& 17.55\,/\,\textbf{1.80} \\
\taskname{Erdos\_Overlap} $\downarrow$   
& 0.38866\,/\,\textbf{0.38093} 
& 800\,/\,\textbf{211.6} 
& 12.74\,/\,\textbf{4.95} \\
\taskname{Matmul\_SwishGN} $\uparrow$     
& 2.55034\,/\,\textbf{3.02228} 
& 100\,/\,\textbf{26.4} 
& 9.83\,/\,\textbf{1.75} \\
\taskname{Conv3d\_Mish} $\uparrow$      
& 1.87863\,/\,\textbf{2.88576} 
& 100\,/\,\textbf{40.2} 
& 4.83\,/\,\textbf{1.44} \\
\taskname{Conv2d\_BN} $\uparrow$        
& 1.41343\,/\,\textbf{2.09174} 
& 100\,/\,\textbf{33.2} 
& 9.88\,/\,\textbf{2.55} \\
\hline
\end{tabular*}

%% file: sections/additional_experiments.tex
\FloatBarrier
\Needspace{6\baselineskip}
\section{TurboEvolve versus Independent Single-Candidate Calls}
\label{appendix:independent_k}

The Independent-$K$ control is defined in Section~\ref{subsec:ablations}.
The full token budgets and evaluated-program counts are reported below.

\subsection{Matched-token budgets}

\sysname{TurboEvolve}'s token consumption at 800 evaluated programs supplies the reference budgets:
3,626,247 input tokens, 393,339 output tokens, and 4,019,586 total tokens. We match each budget
separately. Table~\ref{tab:independent_tokens} reports objective values, while
Table~\ref{tab:independent_counts} gives the evaluated-program counts needed to interpret them.

\par\medskip
\noindent\begin{minipage}{\linewidth}
\centering
\captionof{table}{Matched-token controls on \taskname{Uncertainty} with Gemini-3-Flash.
Entries are mean $\pm$ standard deviation; lower objective values are better.
Each column applies a separate token budget.}
\label{tab:independent_tokens}
\par\vspace{\baselineskip}
\begin{tabular*}{\textwidth}{@{\extracolsep{\fill}}lccc@{}}
\toprule
Method & Input: 3,626,247 & Output: 393,339 & Total: 4,019,586\\
\midrule
\sysname{TurboEvolve} & $\mathbf{0.329957}\mathbf{\pm0.002858}$ & $\mathbf{0.329957}\mathbf{\pm0.002858}$ & $\mathbf{0.329957}\mathbf{\pm0.002858}$\\
\sysname{OpenEvolve} & $0.348475\pm0.012751$ & $0.348368\pm0.012869$ & $0.348475\pm0.012751$\\
Independent-$K$ & $0.353319\pm0.002742$ & $0.350119\pm0.006296$ & $0.353319\pm0.002742$\\
\sysname{PartEvo} & $0.350246\pm0.024267$ & $0.385045\pm0.038397$ & $0.361850\pm0.022181$\\
\sysname{LoongFlow} & $0.374063\pm0.007187$ & $0.374063\pm0.007187$ & $0.374063\pm0.007187$\\
\bottomrule
\end{tabular*}
\end{minipage}
\par

\par\medskip
\noindent\begin{minipage}{\linewidth}
\centering
\captionof{table}{Evaluated-program counts under the token budgets of
Table~\ref{tab:independent_tokens}.}
\label{tab:independent_counts}
\par\vspace{\baselineskip}
\begin{tabular*}{0.68\linewidth}{@{\extracolsep{\fill}}lrr@{}}
\toprule
Matched resource & \sysname{TurboEvolve} & Independent-$K$\\
\midrule
Input tokens & 800 & $169.3\pm15.9$\\
Output tokens & 800 & $204.0\pm22.9$\\
Total tokens & 800 & $173.3\pm14.3$\\
\bottomrule
\end{tabular*}
\end{minipage}
\par

\subsection{Nearly matched evaluated-program horizons}

Table~\ref{tab:independent_horizon} in the main text reports the complete-iteration
comparison specified in Section~\ref{subsec:ablations}.
\sysname{TurboEvolve}'s score differs slightly from 0.329957 in Table~\ref{tab:independent_tokens} because
truncation may stop before 800 programs.

\Needspace{6\baselineskip}
\section{Local Inference with DeepSeek-V4-Flash}
\label{appendix:local_inference}

To examine transfer beyond Gemini APIs and provider token pricing, we evaluate \sysname{TurboEvolve},
\sysname{OpenEvolve}, \sysname{LoongFlow}, and \sysname{PartEvo} with local DeepSeek-V4-Flash inference on four NVIDIA H200 GPUs.
Tables~\ref{tab:local_quality} and~\ref{tab:local_compute} report solution quality,
evaluated-program counts, and both active and reserved GPU hours on \taskname{Erdos\_Overlap}
and \taskname{Conv2d\_BN}. Per-program summaries are reported separately from aggregate totals.

\par\medskip
\noindent\begin{minipage}{\linewidth}
\centering
\captionof{table}{Local DeepSeek-V4-Flash results on four H200 GPUs. Entries are mean $\pm$ standard
deviation. Bold marks the strongest mean objective within each task; these are descriptive
comparisons, not significance tests.}
\label{tab:local_quality}
\par\vspace{\baselineskip}
\begin{tabular*}{0.80\linewidth}{@{\extracolsep{\fill}}lrr@{}}
\toprule
Method & Objective & Evaluated programs\\
\midrule
\multicolumn{3}{c}{\taskname{Erdos\_Overlap} ($\downarrow$)}\\
\midrule
\sysname{TurboEvolve} & \makecell[r]{$\mathbf{0.380951390}$\\$\mathbf{\pm0.000001475}$} & \makecell[r]{$798.5$\\$\pm0.5$}\\
\sysname{OpenEvolve} & \makecell[r]{$0.383684278$\\$\pm0.002500354$} & \makecell[r]{$241.0$\\$\pm32.9$}\\
\sysname{LoongFlow} & \makecell[r]{$0.382213364$\\$\pm0.001534239$} & \makecell[r]{$109.0$\\$\pm13.1$}\\
\sysname{PartEvo} & \makecell[r]{$0.381014212$\\$\pm0.000042971$} & \makecell[r]{$196.3$\\$\pm4.5$}\\
\midrule
\multicolumn{3}{c}{\taskname{Conv2d\_BN} ($\uparrow$)}\\
\midrule
\sysname{TurboEvolve} & \makecell[r]{$1.467937$\\$\pm0.243643$} & \makecell[r]{$100.7$\\$\pm0.9$}\\
\sysname{OpenEvolve} & \makecell[r]{$1.455945$\\$\pm0.252503$} & \makecell[r]{$34.7$\\$\pm1.2$}\\
\sysname{LoongFlow} & \makecell[r]{$\mathbf{1.471607}$\\$\mathbf{\pm0.384386}$} & \makecell[r]{$20.3$\\$\pm1.2$}\\
\sysname{PartEvo} & \makecell[r]{$1.017147$\\$\pm0.017070$} & \makecell[r]{$10.7$\\$\pm3.1$}\\
\bottomrule
\end{tabular*}
\end{minipage}
\par

\sysname{TurboEvolve} uses approximately 58--82\% less active GPU time per evaluated program than the
baselines on \taskname{Erdos\_Overlap}, and 54--82\% less on \taskname{Conv2d\_BN}.
It achieves the lowest mean objective on \taskname{Erdos\_Overlap} and a competitive kernel speedup
(1.467937 versus \sysname{LoongFlow}'s 1.471607). These observations support transfer of the per-program
efficiency advantage beyond Gemini pricing. They do not imply the lowest total GPU consumption
or the lowest reserved GPU time per program in every task: for example, \sysname{OpenEvolve} has lower
reserved GPU-h/program on \taskname{Conv2d\_BN}.

\par\medskip
\noindent\begin{minipage}{\linewidth}
\centering
\captionof{table}{Local-inference GPU resource measurements (mean $\pm$ standard deviation).
Active and reserved GPU hours are reported separately. Total resource use and resource use per
evaluated program answer different questions; the methods evaluate different numbers of programs.}
\label{tab:local_compute}
\par\vspace{\baselineskip}
\begin{minipage}[t]{0.48\linewidth}
\vspace{0pt}\centering
\setlength{\tabcolsep}{3pt}
\textbf{(A) Total resource use}\par\smallskip
\begin{tabular*}{\linewidth}{@{\extracolsep{\fill}}lrr@{}}
\toprule
Method & \makecell{Active\\GPU-h\\\strut} & \makecell{Reserved\\GPU-h\\\strut}\\
\midrule
\multicolumn{3}{c}{\taskname{Erdos\_Overlap}}\\
\midrule
\sysname{TurboEvolve} & \makecell[r]{$8.225$\\$\pm0.164$} & \makecell[r]{$33.314$\\$\pm0.485$}\\
\sysname{OpenEvolve} & \makecell[r]{$5.899$\\$\pm0.599$} & \makecell[r]{$16.199$\\$\pm3.813$}\\
\sysname{LoongFlow} & \makecell[r]{$6.179$\\$\pm0.404$} & \makecell[r]{$23.395$\\$\pm12.828$}\\
\sysname{PartEvo} & \makecell[r]{$10.300$\\$\pm0.963$} & \makecell[r]{$17.852$\\$\pm0.096$}\\
\midrule
\multicolumn{3}{c}{\taskname{Conv2d\_BN}}\\
\midrule
\sysname{TurboEvolve} & \makecell[r]{$1.191$\\$\pm0.092$} & \makecell[r]{$4.258$\\$\pm1.143$}\\
\sysname{OpenEvolve} & \makecell[r]{$0.891$\\$\pm0.140$} & \makecell[r]{$1.229$\\$\pm0.229$}\\
\sysname{LoongFlow} & \makecell[r]{$0.645$\\$\pm0.106$} & \makecell[r]{$0.865$\\$\pm0.193$}\\
\sysname{PartEvo} & \makecell[r]{$0.654$\\$\pm0.104$} & \makecell[r]{$0.730$\\$\pm0.102$}\\
\bottomrule
\end{tabular*}
\end{minipage}\hfill
\begin{minipage}[t]{0.48\linewidth}
\vspace{0pt}\centering
\setlength{\tabcolsep}{3pt}
\textbf{(B) Resource use per evaluated program}\par\smallskip
\begin{tabular*}{\linewidth}{@{\extracolsep{\fill}}lrr@{}}
\toprule
Method & \makecell{Active\\GPU-h/\\program} & \makecell{Reserved\\GPU-h/\\program}\\
\midrule
\multicolumn{3}{c}{\taskname{Erdos\_Overlap}}\\
\midrule
\sysname{TurboEvolve} & \makecell[r]{$0.010301$\\$\pm0.000199$} & \makecell[r]{$0.041721$\\$\pm0.000582$}\\
\sysname{OpenEvolve} & \makecell[r]{$0.024617$\\$\pm0.001202$} & \makecell[r]{$0.066246$\\$\pm0.007115$}\\
\sysname{LoongFlow} & \makecell[r]{$0.057363$\\$\pm0.006772$} & \makecell[r]{$0.216607$\\$\pm0.117452$}\\
\sysname{PartEvo} & \makecell[r]{$0.052410$\\$\pm0.004047$} & \makecell[r]{$0.090985$\\$\pm0.002513$}\\
\midrule
\multicolumn{3}{c}{\taskname{Conv2d\_BN}}\\
\midrule
\sysname{TurboEvolve} & \makecell[r]{$0.011833$\\$\pm0.000928$} & \makecell[r]{$0.042342$\\$\pm0.011558$}\\
\sysname{OpenEvolve} & \makecell[r]{$0.025664$\\$\pm0.003491$} & \makecell[r]{$0.035391$\\$\pm0.005908$}\\
\sysname{LoongFlow} & \makecell[r]{$0.031853$\\$\pm0.005538$} & \makecell[r]{$0.042680$\\$\pm0.009938$}\\
\sysname{PartEvo} & \makecell[r]{$0.064132$\\$\pm0.012861$} & \makecell[r]{$0.071842$\\$\pm0.014328$}\\
\bottomrule
\end{tabular*}
\end{minipage}
\end{minipage}
\par

\Needspace{6\baselineskip}
\section{Resource Accounting beyond API Cost}
\label{appendix:resources}

\subsection{Caching and shared-context amortization}

Provider-side prompt/prefix caching is incorporated into the API costs reported in the controlled
comparisons, using the same accounting rule for all methods within each backend block.
The single-candidate baselines therefore receive the caching benefits available under the provider's
billing policy. A shared-context accounting analysis estimates that joint generation reduces repeated
input-token usage by 60.1--85.2\% (median 76.5\%) relative to generating the retained candidates in
separate calls. This is an input-token amortization measure, not a claim about total API cost or
end-to-end runtime savings.

\subsection{Matched-resource quality}

We additionally compare quality while separately controlling input tokens, output tokens,
cached-input usage, cache-aware API cost, evaluator compute, and active wall-clock time.
Across the 16 task--backend settings, 64 resource comparisons have sufficient baseline resource
coverage. \sysname{TurboEvolve} ranks first in 55 of these and second in the remaining nine.
This summary applies to the covered comparisons rather than all $16\times6$ potential combinations.
Table~\ref{tab:resource_quality} reports the representative \taskname{Uncertainty} results
under Gemini-3-Flash. These resource controls are a separate analysis from the Independent-$K$
experiment in Appendix~\ref{appendix:independent_k}; their values should not be interchanged.

\par\medskip
\noindent\begin{minipage}{\linewidth}
\centering
\captionof{table}{Representative resource-matched quality on \taskname{Uncertainty} with Gemini-3-Flash.
Entries are mean $\pm$ standard deviation; lower is better. Each row independently matches one
resource. This table is distinct from the token-budget experiment in Table~\ref{tab:independent_tokens}.}
\label{tab:resource_quality}
\par\vspace{\baselineskip}
\begin{tabular*}{\textwidth}{@{\extracolsep{\fill}}lcccc@{}}
\toprule
Matched resource & \sysname{TurboEvolve} & \sysname{OpenEvolve} & \sysname{LoongFlow} & \sysname{PartEvo}\\
\midrule
Input tokens & \makecell[r]{$\mathbf{0.338952}$\\$\mathbf{\pm0.013404}$} & \makecell[r]{$0.361739$\\$\pm0.017780$} & \makecell[r]{$0.374063$\\$\pm0.007187$} & \makecell[r]{$0.367575$\\$\pm0.018286$}\\
Output tokens & \makecell[r]{$\mathbf{0.338211}$\\$\mathbf{\pm0.013874}$} & \makecell[r]{$0.360487$\\$\pm0.019143$} & \makecell[r]{$0.374063$\\$\pm0.007187$} & \makecell[r]{$0.361777$\\$\pm0.022264$}\\
Cached-input usage & \makecell[r]{$\mathbf{0.345516}$\\$\mathbf{\pm0.024338}$} & \makecell[r]{$0.361187$\\$\pm0.018641$} & \makecell[r]{$0.379927$\\$\pm0.007187$} & \makecell[r]{$0.368904$\\$\pm0.015781$}\\
API cost & \makecell[r]{$\mathbf{0.338211}$\\$\mathbf{\pm0.013874}$} & \makecell[r]{$0.360849$\\$\pm0.018612$} & \makecell[r]{$0.374063$\\$\pm0.007187$} & \makecell[r]{$0.361777$\\$\pm0.022264$}\\
Evaluator compute & \makecell[r]{$\mathbf{0.338662}$\\$\mathbf{\pm0.013573}$} & \makecell[r]{$0.360504$\\$\pm0.019125$} & \makecell[r]{$0.372746$\\$\pm0.009043$} & \makecell[r]{$0.367575$\\$\pm0.018286$}\\
Active wall time & \makecell[r]{$\mathbf{0.338662}$\\$\mathbf{\pm0.013573}$} & \makecell[r]{$0.360504$\\$\pm0.019125$} & \makecell[r]{$0.374063$\\$\pm0.007187$} & \makecell[r]{$0.361777$\\$\pm0.022264$}\\
\bottomrule
\end{tabular*}
\end{minipage}
\par

\sysname{TurboEvolve} has the lowest mean objective under all six resource controls in this example.
The improvement over the strongest baseline ranges from approximately 4.3\% to 6.3\%, depending
on the resource. These controls support robustness to alternative accounting choices in the
examined setting, while the unequal coverage limits the scope of the aggregate ranking.

\subsection{Evaluator compute and active wall time}

Table~\ref{tab:evaluator_wall} reports an explicit breakdown for \taskname{CirclePack\_26} with
Gemini-3-Flash. At comparable API cost, \sysname{TurboEvolve} evaluates $797.4$ programs on average,
versus $72.6$--$233.6$ for the baselines. Its total evaluator time or total wall time therefore
need not be the smallest. Resource use per program provides a different view: \sysname{TurboEvolve}
requires 2.87 evaluator seconds and 7.78 active-wall seconds per evaluated program in this example.

\par\medskip
\noindent\begin{minipage}{\linewidth}
\centering
\captionof{table}{Evaluator and active-wall-time breakdown on \taskname{CirclePack\_26} with Gemini-3-Flash
(mean $\pm$ standard deviation). Per-program measurements are reported separately from aggregate totals.}
\label{tab:evaluator_wall}
\par\vspace{\baselineskip}
\begin{tabular*}{\textwidth}{@{\extracolsep{\fill}}lrrrr@{}}
\toprule
Measure & \sysname{TurboEvolve} & \sysname{OpenEvolve} & \sysname{LoongFlow} & \sysname{PartEvo}\\
\midrule
API cost (\$) & $1.9598\pm0.3079$ & $1.9501\pm0.0050$ & $1.8559\pm0.1055$ & $1.9555\pm0.0038$\\
Evaluated programs & $797.4\pm2.1$ & $107.0\pm15.9$ & $72.6\pm30.3$ & $233.6\pm18.0$\\
Evaluator compute (h) & $0.63\pm0.16$ & $0.18\pm0.08$ & $1.03\pm0.69$ & $0.20\pm0.05$\\
Active wall time (h) & $1.72\pm0.71$ & $0.90\pm0.35$ & $3.72\pm0.83$ & $1.61\pm0.31$\\
Evaluator s/program & $2.87\pm0.71$ & $6.01\pm2.09$ & $49.33\pm21.03$ & $3.07\pm0.62$\\
Active wall s/program & $7.78\pm3.21$ & $29.93\pm9.02$ & $211.65\pm80.56$ & $24.62\pm3.53$\\
\bottomrule
\end{tabular*}
\end{minipage}
\par

Across all 16 task--backend settings, the aggregate per-program ranking places \sysname{TurboEvolve} first
for API cost and active wall time, and second for evaluator compute. This distinction avoids
equating improved throughput with universally lower total compute expenditure.